\documentclass{article}

\usepackage{arxiv}

\usepackage{amsmath}
\usepackage{amssymb}
\usepackage{mathtools}
\usepackage{amsthm}

\usepackage[utf8]{inputenc} % allow utf-8 input
\usepackage[T1]{fontenc}    % use 8-bit T1 fonts
\usepackage{hyperref}       % hyperlinks
\usepackage{url}            % simple URL typesetting
\usepackage{booktabs}       % professional-quality tables
\usepackage{amsfonts}       % blackboard math symbols
\usepackage{nicefrac}       % compact symbols for 1/2, etc.
\usepackage{microtype}      % microtypography
\usepackage[capitalize,noabbrev]{cleveref}       % smart cross-referencing
\usepackage{lipsum}         % Can be removed after putting your text content
\usepackage{graphicx}
\usepackage{doi}
\usepackage[dvipsnames]{xcolor}
\usepackage{multirow}
\usepackage{subcaption}
\usepackage{algorithm}
\usepackage{algorithmic}

\usepackage{natbib}
\renewcommand{\cite}[1]{\citep{#1}}
\setcitestyle{authoryear,round,citesep={;},aysep={,},yysep={;}}

\title{PIGEON: Predicting Image Geolocations}

\date{}

\author{Lukas Haas\\
	Department of Computer Science\\
	Stanford University\\
	\texttt{lukashaas@cs.stanford.edu} \\
	\And
        Michal Skreta\\
	Department of Computer Science\\
	Stanford University\\
	\texttt{michal.skreta@stanford.edu} \\
        \And
	Silas Alberti\\
	Department of Electrical Engineering\\
	Stanford University\\
	\texttt{salberti@stanford.edu} \\
        \And
        Chelsea Finn\\
	Department of Computer Science\\
	Stanford University\\
	\texttt{cbfinn@cs.stanford.edu} \\
}

\renewcommand{\headeright}{}
\renewcommand{\undertitle}{Preprint}
\renewcommand{\shorttitle}{PIGEON: Predicting Image Geolocations}

\hypersetup{
pdftitle={PIGEON Preprint},
pdfsubject={},
pdfauthor={Lukas Haas, Silas Alberti, Michal Skreta, Chelsea Finn},
pdfkeywords={Image Geolocalization, Visual Geolocation Recognition, CLIP, Computer Vision, Multi-Modal, Multi-Task, Geoguessr},
}

\begin{document}
\maketitle

\begin{abstract}
Planet-scale image geolocalization remains a challenging problem due to the diversity of images originating from anywhere in the world. Although approaches based on vision transformers have made significant progress in geolocalization accuracy, success in prior literature is constrained to narrow distributions of images of landmarks, and performance has not generalized to unseen places. We present a new geolocalization system that combines semantic geocell creation, multi-task contrastive pretraining, and a novel loss function. Additionally, our work is the first to perform retrieval over location clusters for guess refinements. We train two models for evaluations on street-level data and general-purpose image geolocalization; the first model, PIGEON, is trained on data from the game of GeoGuessr and is capable of placing over 40\% of its guesses within 25 kilometers of the target location globally. We also develop a bot and deploy PIGEON in a blind experiment against humans, ranking in the top 0.01\% of players. We further challenge one of the world's foremost professional GeoGuessr players to a series of six matches with millions of viewers, winning all six games. Our second model, PIGEOTTO, differs in that it is trained on a dataset of images from Flickr and Wikipedia, achieving state-of-the-art results on a wide range of image geolocalization benchmarks, outperforming the previous SOTA by up to 7.7 percentage points on the city accuracy level and up to 38.8 percentage points on the country level. Our findings suggest that PIGEOTTO is the first image geolocalization model that effectively generalizes to unseen places and that our approach can pave the way for highly accurate, planet-scale image geolocalization systems. Our code is available on GitHub.\footnote{\url{https://github.com/LukasHaas/PIGEON}.}
\end{abstract}

\vspace{1em}

% keywords can be removed
\keywords{Image Geolocalization \and Visual Place Recognition \and Photo Geolocalization \and Computer Vision \and Semantic Geocells \and Multi-Task Pretraining \and Haversine \and Location Refinement \and Clustering \and Voronoi \and Multi-Modal \and Geoguessr}

\section{Introduction}
\label{sec:intro}

%-------------------------------------------------------------------------
% Figure: Inference Pipeline
%-------------------------------------------------------------------------
% \begin{figure*}[!htbp]
% \centering
% \includegraphics[width=0.6\linewidth]{figures/inference_pipeline.png}
% \caption{A graphic of our inference pipeline will appear here. \textcolor{red}{This is our teaser figure.}}
% \label{fig:inference_pipeline}
% \end{figure*}

\begin{figure*}[tbhp]
\vskip -0.15in
\centering
\centerline{\includegraphics[width=1.0 \linewidth]{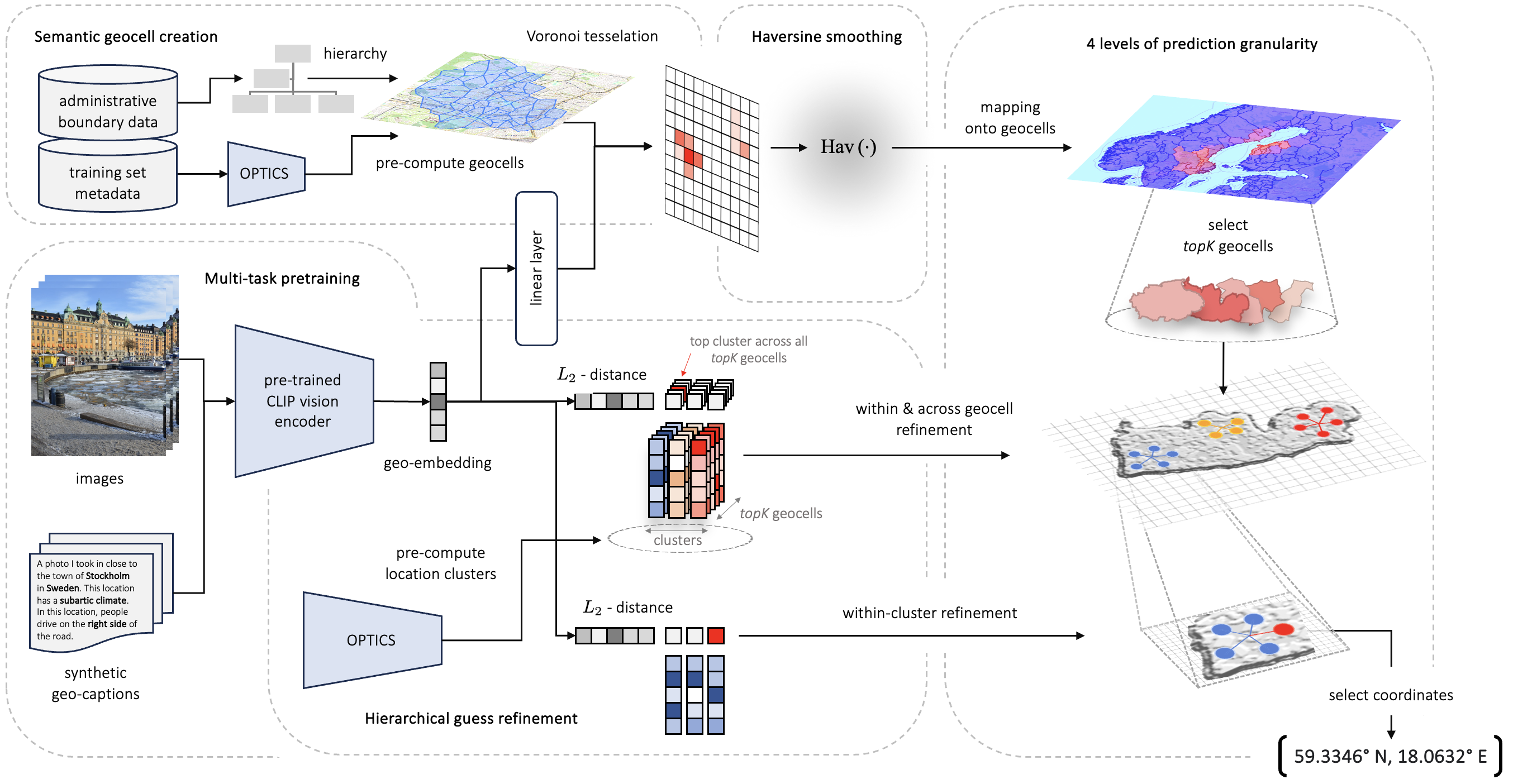}}
\caption{\textbf{Prediction pipeline and main contributions of PIGEON}. Administrative boundary and training set metadata are hierarchically ranked, clustered, and Voronoi tessellated to create semantic geocells. The geocell labels are then used to create continuous labels via haversine smoothing. Additionally, we pretrain CLIP via geographic synthetic captions in a multi-task setting. The pretrained CLIP model together with an OPTICS clustering model are employed to generate location cluster representations. During inference, an image embedding is computed and first passed to a linear layer to create geocell predictions and to identify the \textit{topK} geocell candidates. The embedding is also used in our refinement pipeline to refine predictions within and across geocells. This is achieved by minimizing the embedding $L_2$-distance between the inference image embedding and all location cluster representations across the \textit{topK} geocells. Finally, predictions are refined within the top identified cluster to generate geographic coordinates as outputs.}
\label{fig:pipeline}
% \vskip -0.2in
\end{figure*}

%-------------------------------------------------------------------------

The online game \href{https://www.geoguessr.com/}{GeoGuessr} has recently reached 65 million players~\cite{lucas_2023}, attracting a worldwide crowd of users trying to solve a single problem: given a Street View image taken somewhere in the world, identify its location. The problem of uncovering geographical coordinates from visual data is more formally known in computer vision as image geolocalization, and, just like the game of GeoGuessr, remains notoriously challenging. The scale and diversity of our planet, seasonal appearance disturbance, and climate change impacts are some among the many reasons why image geolocalization remains an unsolved problem.

Over the past decade, researchers have advanced the field by casting image geolocalization as a classification task~\cite{weyand_et_al_2016}, developing hierarchical approaches to problem modeling~\cite{mueller_budack_et_al_2018, pramanick_et_al_2022, clark_et_al_2023}, as well as leveraging vision transformers~\cite{pramanick_et_al_2022, clark_et_al_2023} and contrastive pretraining~\cite{luo_et_al_2022}. Yet despite this progress, the most capable models have been highly dependent on distributional alignments between training and testing data, failing to generalize to more diverse datasets that predominantly include unseen locations~\cite{clark_et_al_2023}.

In this work, we present a two-pronged multi-task modeling approach that both exhibits world-leading performance in the game of GeoGuessr and achieves state-of-the-art performance on a wide range of image geolocalization benchmark datasets. First, we present \textbf{PIGEON}, a model trained exclusively on planet-scale Street View data, taking a four-image panorama as input. PIGEON is the first computer vision model to reliably beat the most experienced players in the game GeoGuessr, comfortably ranking within the top 0.01\% of players while also beating one of the world's best professional players in six out of six games with millions of viewers. Our model achieves impressive image geolocalization results on outdoor street-level images globally, placing 40.4\% of its geographic coordinate predictions within a 25-kilometer radius of the correct location.

Subsequently, we evolve our model to \textbf{PIGEOTTO} which differs from PIGEON in that it takes a single image as input and is trained on a larger, highly diverse dataset of over 4 million photos derived from Flickr and Wikipedia and no Street View data. PIGEOTTO achieves state-of-the-art results across a wide range of benchmark datasets, including IM2GPS \cite{hays_and_efros_2008}, IM2GPS3k \cite{vo_et_al_2017}, YFCC4k \cite{vo_et_al_2017}, YFCC26k \cite{mueller_budack_et_al_2018}, and GWS15k \cite{clark_et_al_2023}. The model slashes the median distance error roughly in half on three benchmark datasets and more than five times reduces the median error on GWS15k which includes images from predominantly unseen locations. PIGEOTTO is the first model that is robust to location and image distribution shifts by picking up general locational cues in images as evidenced by the often double-digit percentage-point increase in performance on larger evaluation radii. By performing well on out-of-distribution datasets, PIGEOTTO closes a major gap in prior literature that is essential for solving the problem of image geolocalization.

As PIGEON and PIGEOTTO only differ in the training data and hyperparameter settings, the efficacy of our approach has important implications for planet-scale image geolocalization. Our contributions of semantic geocells, multi-task contrastive pretraining, a new loss function, and downstream guess refinement all contribute to minimizing distance errors, as shown in our ablation studies in~\Cref{sec:experiments}. Still, it is important that future research addresses the safety of image geolocalization technologies, ensuring responsible progress in developing computer vision systems.

\section{Related work}
\label{sec:related_work}

%-------------------------------------------------------------------------
\subsection{Image geolocalization problem setting}
\label{sec:problem_setup}

Image geolocalization refers to the problem of mapping an image to coordinates that identify where it was taken. This problem, especially if planet-scale, remains a very challenging area of computer vision. Not only does a global problem formulation render the problem intractable, but accurate image geolocalization is also difficult due to changes in daytime, weather, seasons, time, illumination, climate, traffic, viewing angle, and many more factors.

The first modern attempt at planet-scale image geolocalization is attributed to IM2GPS (2008)~\cite{hays_and_efros_2008}, a retrieval-based approach based on hand-crafted features. Dependence on nearest-neighbor retrieval methods~\cite{zamir_and_shah_2014} using hand-crafted visual features~\cite{crandall_et_al_2009} meant that an enormous database of reference images would be necessary for accurate planet-scale geolocalization, which is infeasible. Consequently, subsequent work decided to restrict the geographic scope, focusing instead on specific cities~\cite{wu_and_huang_2022} like Orlando and Pittsburgh~\cite{zamir_and_shah_2010} or San Francisco~\cite{berton_et_al_2022}; specific countries like the United States~\cite{suresh_et_al_2018}; and even mountain ranges~\cite{baatz_2012, saurer_et_al_2016, tomesek_et_al_2022}, deserts~\cite{tzeng_et_al_2013}, and beaches~\cite{cao_et_al_2012}.

%-------------------------------------------------------------------------
\subsection{Vision transformers and multi-task learning}
\label{sec:problem_setup}

With the advent of deep learning, methods in image geolocalization shifted from hand-crafted features to end-to-end learning~\cite{masone_and_caputo_2021}. In 2016, Google released the PlaNet~\cite{weyand_et_al_2016} paper that first applied convolutional neural networks (CNNs)~\cite{krizhevsky_et_al_2012} to geolocalization. It also first cast the problem as a classification task across ``geocells" as a response to research demonstrating that it was difficult for deep learning models to directly predict geographic coordinates via regression~\cite{deBrebisson_et_al_2015, theiner_et_al_2021}. This was due to the subtleties in geographic data distributions and the complex interdependence between latitudes and longitudes. The improvements realized with deep learning led researchers to revisit IM2GPS~\cite{vo_et_al_2017}, apply CNNs to massive datasets of mobile images~\cite{howard_et_al_2017}, and deploy their models in the game of GeoGuessr against human players~\cite{suresh_et_al_2018, luo_et_al_2022}. Prior literature has also combined classification and retrieval approaches~\cite{kordopatis_zilos_et_al_2021}; our work modernizes this approach via a hierarchical retrieval mechanism over location clusters, equivalent to prototypical networks~\cite{snell_2017} with fixed parameters. 

Following the success of transformers~\cite{vaswani_et_al_2017} in natural language processing, the transformer architecture found its application in computer vision. Pretrained vision transformers (ViT)~\cite{kolesnikov_et_al_2021} and multi-modal derivatives such as OpenAI's CLIP \cite{radford_et_al_2021} and GPT-4V~\cite{openai_2023} have successfully been deployed to image geolocalization ~\cite{agarwal_et_al_2021, pramanick_et_al_2022, wu_and_huang_2022, luo_et_al_2022, zhu_et_al_2022, openai_2023}. Our approach is novel in that in pretrains CLIP specifically for the task of image geolocalization in a multi-task fashion via auxiliary geographic, demographic, and climate data. Auxiliary data had previously been shown to aid in image geolocalization~\cite{hays_and_efros_2008, pramanick_et_al_2022}, but our work is the first to use auxiliary data for contrastive pretraining, retaining CLIP's exceptional in-domain generalized zero-shot capabilities that are critical for geolocalization performance~\cite{haas_et_al_2023}.

% %-------------------------------------------------------------------------
% \subsection{Multi-task learning for image geolocalization}
% \label{sec:multi_task_learning}

% Multi-task approaches have been found to improve the performance of primary tasks by using complementary tasks~\cite{ranjan_et_al_2016}, with certain types of task being more beneficial for the main task than others~\cite{bingel_and_sogaard_2017}. This, coupled with the fact that auxiliary information was found to be a vital pre-processing step for image geolocalization~\cite{hays_and_efros_2008, pramanick_et_al_2022}, pointed to the potential of multi-task learning to significantly accelerated the field of image geolocalization.

%-------------------------------------------------------------------------
\subsection{Geocell partitioning}
\label{sec:geocell_partitioning_approaches}

With image geolocalization framed as a classification problem, the chosen method of partitioning the world into geographical classes, or ``geocells``, can have an enormous effect on downstream performance. Previous approaches rely on geocells that are either plainly rectangular, rectangular while respecting the curvature of the Earth and being roughly balanced in class size~\cite{mueller_budack_et_al_2018} (as is the case of Google's S2 library\footnote{\url{https://code.google.com/archive/p/s2-geometry-library}.}), or geocells that are effectively arbitrary as a result of combinatorial partitioning, initializing cells randomly but adjusting their \textit{shapes} based on the training dataset distribution~\cite{seo_et_al_2018}. Hierarchical approaches to geocell creation like in individual scene networks (ISNs)~\cite{mueller_budack_et_al_2018, theiner_et_al_2021} can help preserve semantic information and exploit the hierarchical knowledge at different geospatial resolutions, for instance by categorizing the geocells at the city, region, and country levels.

While the semantic construction of geocells has been found to be of high importance to image geolocalization~\cite{theiner_et_al_2021}, even recently published papers continue to use the S2 library~\cite{kordopatis_zilos_et_al_2021, pramanick_et_al_2022, clark_et_al_2023}. One of the possible reasons for this design choice is that for larger datasets, even the most granular semantic geocells contain too many data points, causing the classification problem to be very imbalanced. Our work addresses this limitation with a novel semantic geocell creation method, combining hierarchical approaches with clustering based on the training data distribution and Voronoi tesselation as the missing link between the two. For the first time, our approach renders semantic geocells useful for any dataset size and geographic distribution.

%-------------------------------------------------------------------------
\subsection{Additional work}
\label{sec:additional_work}

Other notable academic work cites the efficacy of cross-view image geolocalization, especially for rural regions with sparse, ground-level geo-tagged photos. Cross-view approaches can combine land cover attributes and ground-level and overhead imagery to increase robustness through transfer learning~\cite{lin_et_al_2013, yang_et_al_2021, zhu_et_al_2022}. Using land maps in particular is an important avenue for future research; in our work, however, we aim to demonstrate our models' performance relying solely on ground-level images from diverse settings.

%-------------------------------------------------------------------------

\section{Predicting image geolocations}
\label{sec:technical_approach}
Our image geolocalization system consists of both parametric and non-parametric components. This section first explains our data pre-processing pipeline and then walks through how we frame geolocalization as a distance-aware classification problem. We then delineate our pretraining and training stages, and finally describe how we refine location predictions to improve street-level guess performance.

\subsection{Geocell creation}
\label{sec:geocell_creation}
Contemporary methods all frame image geolocalization as a classification exercise, relying on geocells to discretize the Earth's surface into a set number of classes. Our work experiments with two types of geocell creation methods.

\paragraph{Naive geocells.} We first employ naive, rectangular geocells inspired by the S2 library, subdividing every geocell until roughly balanced class sizes are reached. In contrast to S2 partitioning, our rectangular geocells are not of equal geographic size, creating even more balanced classes.
%Our naive geocell creation algorithm works as follows: a single, large rectangle is initialized, covering all training data locations. Then, in a sequence of steps, the rectangle is divided into two smaller rectangles along its longest side, only dividing a rectangle further if the two resulting rectangles contain a minimum number of points. 

\paragraph{Semantic geocells.} One limitation of the S2 library and our naive geocells is that the geocell boundaries are completely arbitrary and thus meaningless in the context of image geolocalization. Ideally, each geocell should capture the distinctive characteristics of its enclosed geographic area. Political and administrative boundaries serve this purpose well as they often not only capture country or region-specific information (i.e. road markings and street signs) but also follow natural boundaries, such as the flow of rivers and mountain ranges which encode geological information.

\begin{figure}[!htbp]
\centering
    \begin{subfigure}[b]{0.48\textwidth}
        \centering
        \includegraphics[width=\textwidth]{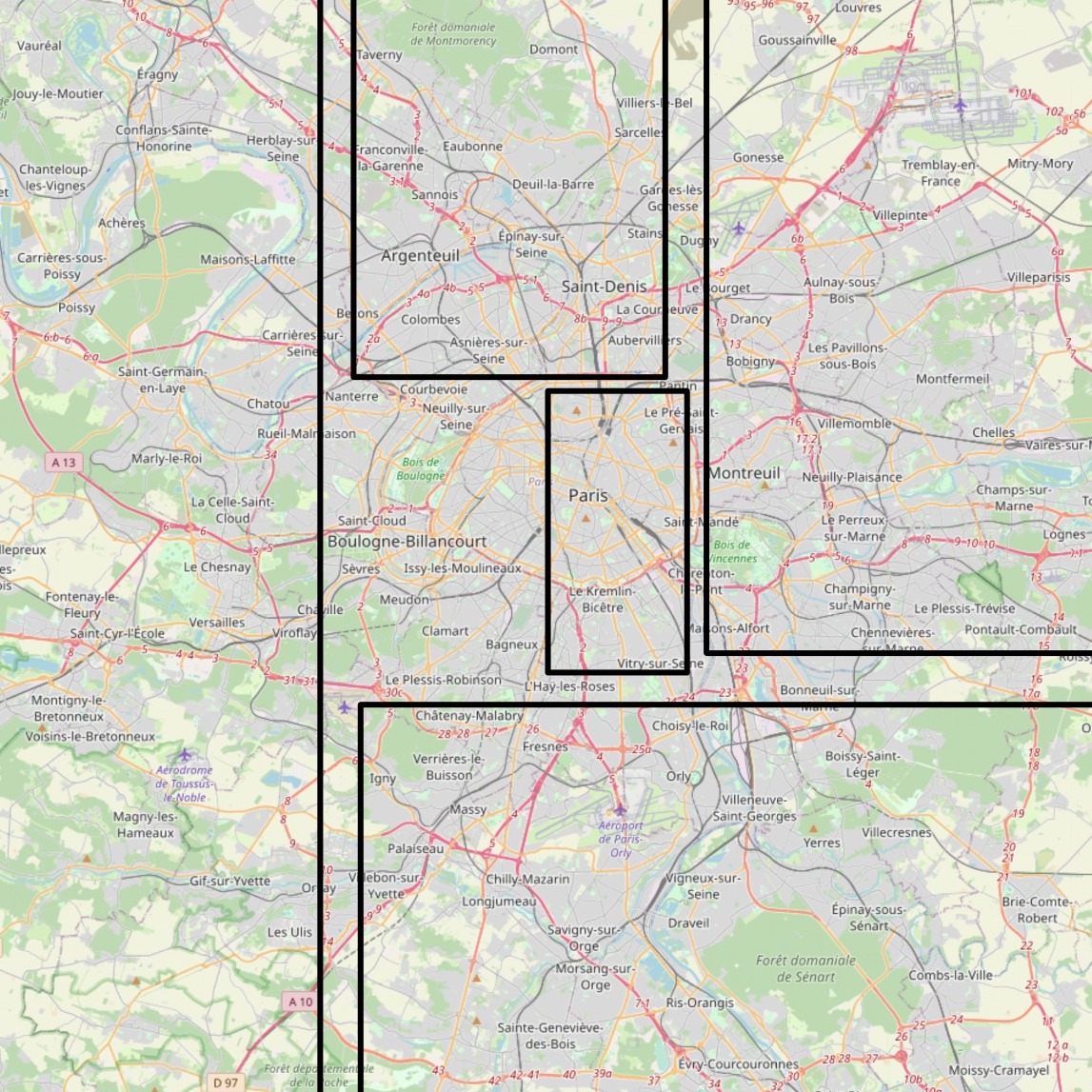}
        \caption{With naive, rectangular geocells.}
        \label{fig:donut_rectangular}
    \end{subfigure}
    \hfill % This will add horizontal space between the subfigures
    \begin{subfigure}[b]{0.48\textwidth}
        \centering
        \includegraphics[width=\textwidth]{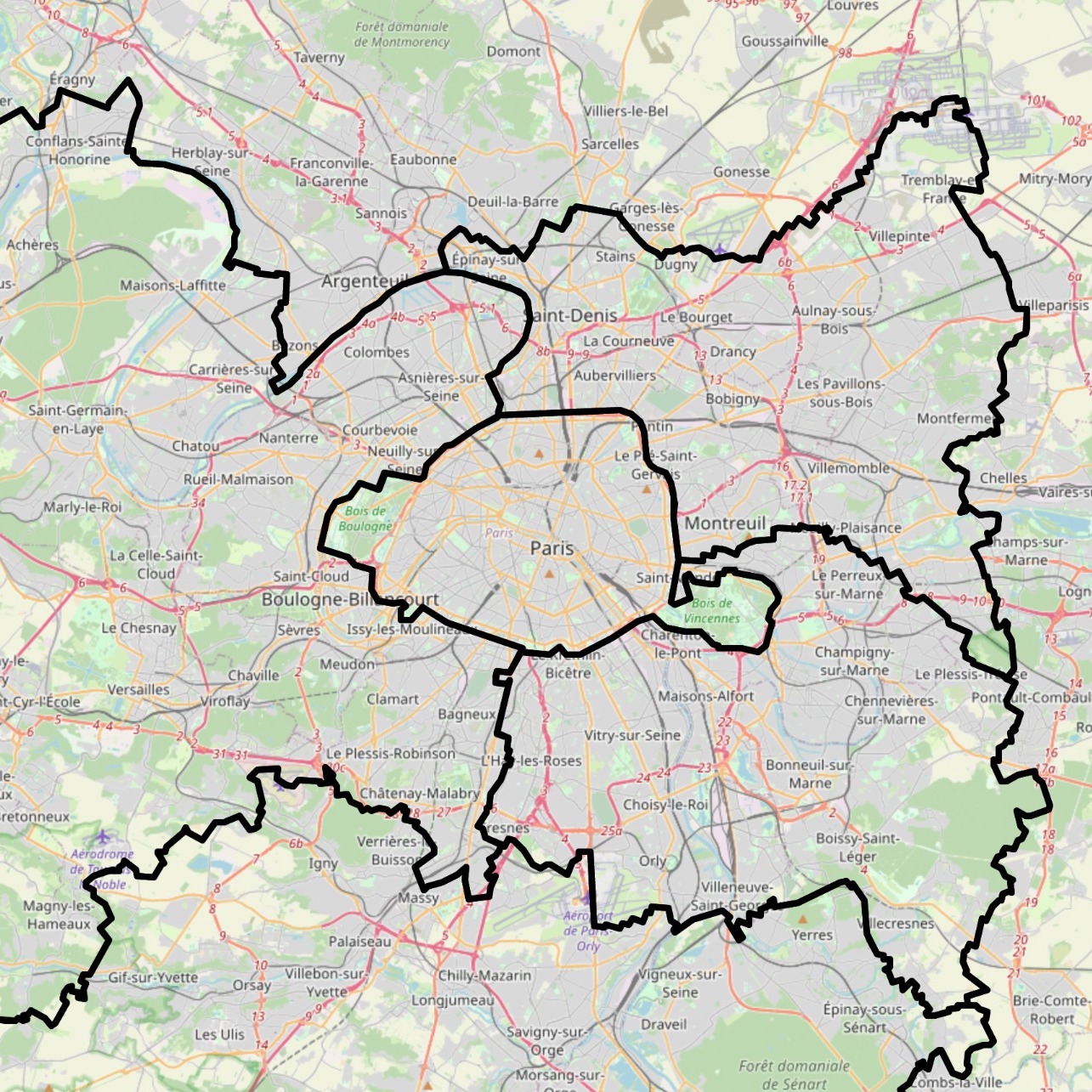}
        \caption{With our semantic geocells.}
        \label{fig:donut_semantic}
    \end{subfigure}
    \vspace{5pt} % Adjust the vertical space here
    \caption{Geocell specifications around Paris, France.}
    \label{fig:donut}
\end{figure}

Similar to~\citet{theiner_et_al_2021}, we rely on planet-scale open-source administrative data for our semantic geocell design, drawing on non-overlapping political shapefiles of three levels of administrative boundaries (country, admin 1, and admin 2 levels) obtained from~\citet{gadm_2022}. Starting at the most granular level (admin 2), our algorithm merges adjacent admin 2 level polygons such that each geocell contains at least a minimum number of training samples. Our method attempts to preserve the hierarchy given by admin 1 level boundaries, never merges cells across country borders (defined by distinct ISO country codes) and, in contrast to \citet{theiner_et_al_2021}, allows for more granular hierarchies. \Cref{fig:donut} shows an example of our semantic geocell design preserving the semantics of urban and surrounding Paris.

%-------------------------------------------------------------------------
% Figure: Donut
%-------------------------------------------------------------------------

\paragraph{OPTICS clustering \& Voronoi tessellation.} We further address a major limitation in the semantic geocell design of \citet{theiner_et_al_2021} which is that some admin 2 areas are not fine-grained enough to result in a balanced classification dataset. This is especially the case for large training datasets where the number of training examples for a single, urban admin 2 area might greatly exceed the minimum class size, requiring admin 2 areas to be meaningfully split further. An important observation is that the geographic distribution of our training data already gives us an indication of how to meaningfully subdivide our geocells because it clusters around popular places and landmarks. We extract these clusters using the OPTICS clustering algorithm \cite{ankerst_1999}. Finally, we assign all yet unassigned data points to their nearest clusters and employ Voronoi tessellation to define contiguous geocells for every extracted cluster.
% because it performs well in settings with many data points, does not require a pre-defined number of clusters, and allows for the specification of minimum cluster sizes

% , as depicted in \Cref{fig:voronoi_tessellation},

% \begin{figure}[!htbp]
% \centering
% \includegraphics[width=0.49\textwidth]{figures/voronoi_tessellation.jpeg}
% \caption{Voronoi tessellation applied in the process of geocell creation for points of an OPTICS cluster in Vienna, Austria.}
% \label{fig:voronoi_tessellation}
% \end{figure}

% Further implementation details can be found in Appendix \ref{sec:appendix_geocells}.

\subsection{Hierarchical image geolocalization using distance-based label smoothing}
By discretizing the problem of image geolocalization, a trade-off is created between the granularity of geocells and predictive accuracy. More granular geocells enable fine-grained predictions but also result in the classification problem becoming more difficult due to a higher cardinality. Prior literature addresses this problem by generating separate geolocalization predictions across multiple levels of geographic granularity, refining guesses at every subsequent level \cite{mueller_budack_et_al_2018, pramanick_et_al_2022, clark_et_al_2023}. \citet{pramanick_et_al_2022} and \citet{clark_et_al_2023} further propose architectures that share some model parameters between different hierarchy levels, improving geolocalization performance. Surprisingly, all prior work suffers from the same limitation: models figuratively guess in the blind as they do not know which geocells are located next to each other, learning their representations in isolation.

Our approach addresses this major limitation and improves upon prior work by sharing \textit{all} parameters between multiple, implicit levels of geographic hierarchies. We achieve this through a new loss function that relates adjacent geocells to each other, biasing the label based on the haversine distance which calculates the distance between two points on the Earth's surface in kilometers. Given two points, $\textbf{p}_1 = (\lambda_1, \phi_1)$ and $\textbf{p}_2 = (\lambda_2, \phi_2)$ with longitude $\lambda$ and latitude $\phi$, and the earth's radius $r$ in kilometers, we define the haversine distance $\text{Hav}(\textbf{p}_1, \textbf{p}_2)$ as follows:

\begin{equation}\label{eq:haversine_distance}
    \normalsize
        \text{Hav}(\textbf{p}_1, \textbf{p}_2) = 2r \arcsin\left(\sqrt{\sin^2 \left(\frac{\phi_2 - \phi_1}{2}\right) + \cos(\phi_1)\cos(\phi_2)\sin^2 \left(\frac{\lambda_2 - \lambda_1}{2}\right)}\right)
\end{equation}\\

We then ``haversine smooth" the original one-hot geocell classification label using this distance metric according to the following equation for a given sample $n$ and geocell $i$: 

\begin{equation}
    y_{n, i} = \exp\left( - \frac{\text{Hav}(\textbf{g}_i, \textbf{x}_n) - \text{Hav}(\textbf{g}_n, \textbf{x}_n)}{\tau}\right)
    \label{eq:smooth}
\end{equation}\\

where $\textbf{g}_i$ are the centroid coordinates of the geocell polygon of cell $i$, $\textbf{g}_n$ are the centroid coordinates of the true geocell, $\textbf{x}_n$ are the true coordinates of the example for which the label is computed, and $\tau$ is a temperature parameter which is set to 75 for PIGEON and to 65 for PIGEOTTO in our experiments. It is important to note that our ``haversine smoothing" is distinct from classical ``label smoothing" because labels are not decayed using a constant factor but based on both the distance to the correct geocell and the true location. Since for every training example, multiple geocells will have a target $y_{n,i}$ that is significantly larger than zero, our model simultaneously learns to predict the correct geocell as well as an even coarser level of geographic granularity. We design the following loss function based on haversine smoothing for a particular training sample $n$:

\begin{equation}
\normalsize
    \mathcal{L}_n = -\sum_{g_i \in G} \log \left(p_{n, i}\right) \cdot \exp\left(- \frac{\text{Hav}(\textbf{g}_i, \textbf{x}_n) - \text{Hav}(\textbf{g}_n, \textbf{x}_n)}{\tau}\right)
    \label{eq:local_loss}
\end{equation}\\

where $p_{n, i}$ is the probability our model assigns to geocell $i$ for sample $n$. An added benefit of using the loss of~\Cref{eq:local_loss} is that it aids generalization because hierarchy definitions vary across every training sample. Additionally, if a sample lies close to the boundary of two geocells, this fact will be reflected through approximately equal target labels for these two geocells. This is especially helpful for larger, often rural, geocells. Furthermore, because every target label $y_{n,i}$ is now continuous and the difficulty of the classification problem can be freely adjusted using $\tau$, an arbitrary number of geocells can be employed as long as geocells are still contextually meaningful and contain a minimum number of samples. Finally, we observe that our classification loss is now directly based on the distance to the true location $\textbf{x}_n$ of a given sample while circumventing the regression difficulties encountered in prior literature \cite{deBrebisson_et_al_2015, theiner_et_al_2021}.

\begin{figure}[!htbp]
\centering
    \begin{subfigure}[b]{0.48\textwidth}
        \centering
        \includegraphics[width=\textwidth]{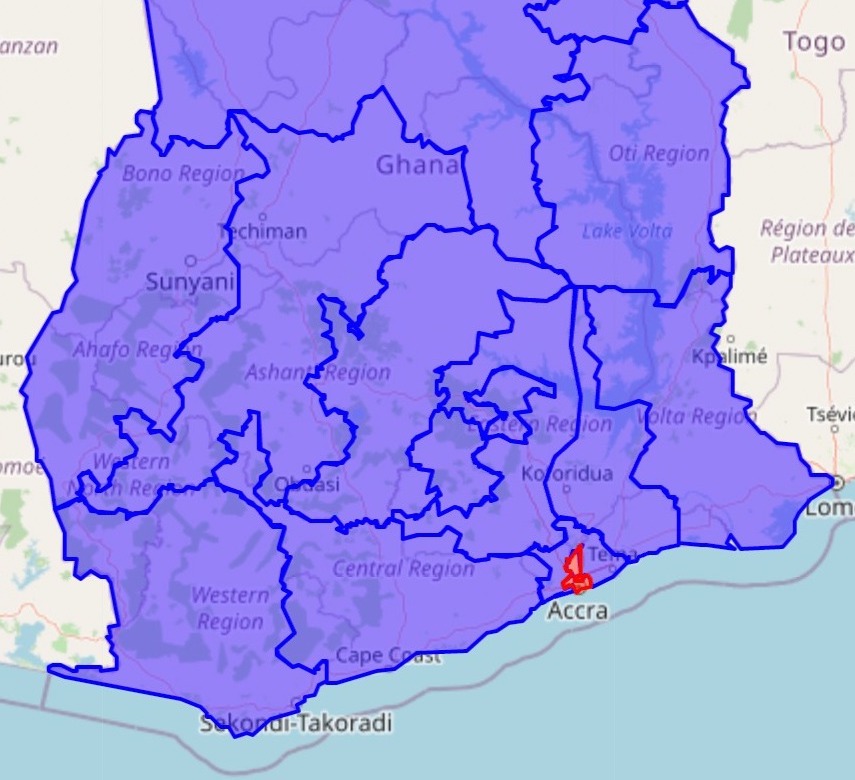}
        \caption{Without haversine smoothing.}
        \label{fig:without_ls}
    \end{subfigure}
    \hfill % This will add horizontal space between the subfigures
    \begin{subfigure}[b]{0.48\textwidth}
        \centering
        \includegraphics[width=\textwidth]{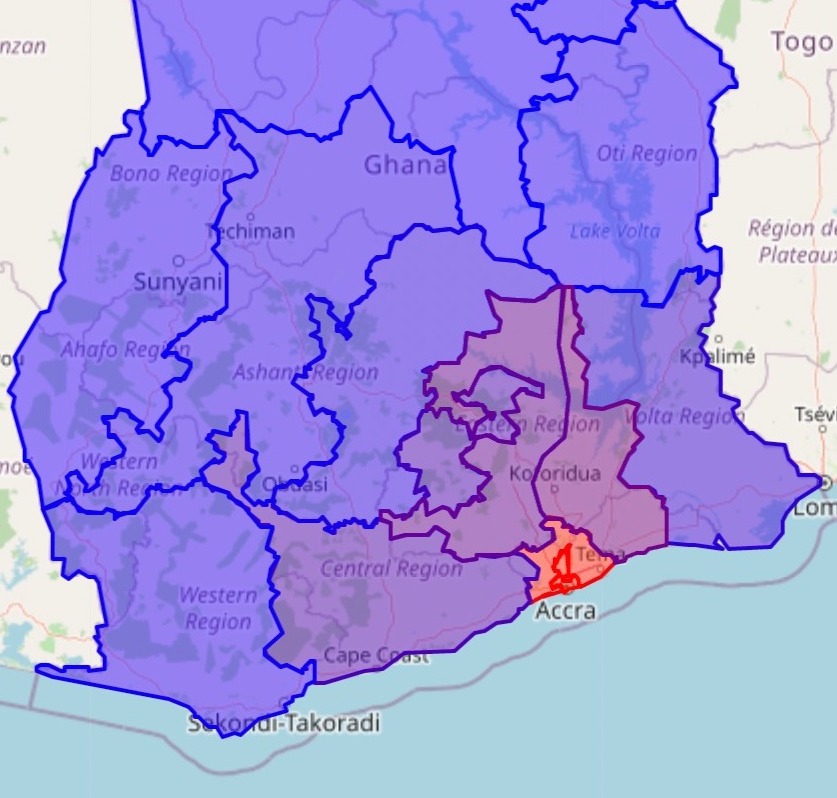}
        \caption{With haversine smoothing.}
        \label{fig:with_ls}
    \end{subfigure}
    \vspace{5pt} % Adjust the vertical space here
    \caption{Impact of applying haversine smoothing over neighboring geocells for a location in Accra, Ghana.}
    \label{fig:label_smoothing}
\end{figure}

\subsection{Contrastive pretraining for geolocalization}
\label{sec:pretraining}

To generate visual representations to then project onto our geocells, our architecture uses OpenAI's CLIP ViT-L/14 336 model as a backbone which is a multi-modal model that was pretrained on a dataset of 400 million images and captions~\cite{radford_et_al_2021}. The reason why we employ CLIP is that it has been shown to perform exceptionally well in generalized zero-shot learning setups \cite{radford_et_al_2021}, which is a desirable property for image geolocalization of both \textit{seen} and \textit{unseen} places.

In our experiments, we add a linear layer on top of CLIP's vision encoder to predict geocells. For model versions with multiple image inputs (i.e. four-image panorama for PIGEON), we average the embeddings of all images. Averaging embeddings resulted in a superior performance compared to combining multiple embeddings via multi-head attention or additional transformer layers.

In \citet{haas_et_al_2023}, the authors demonstrate that continuing the pretraining of CLIP using domain-specific, synthetic captions derived from caption templates improves the generalized zero-shot performance on image geolocalization tasks. We further improve upon their method through the continued pretraining of CLIP in a \textit{multi-task} fashion.

To this end, we augment our training datasets with geographic, climate, and directional auxiliary data, used to create synthetic image captions by sampling caption components from different category templates and concatenating them. For PIGEOTTO, we use caption components based on the location, climate, and traffic direction. Meanwhile, for PIGEON, the Street View metadata allows us to additionally infer compass directions and the season, the latter included to avoid shortcut learning~\cite{geirhos_et_al_2020} (i.e. snow $\rightarrow$ polar latitudes). Examples of caption components include:\vspace{0.7em}

{\footnotesize
\begin{itemize}
    \item \textbf{Location}: ``A photo I took in the region of Gauteng in South Africa."
    \item \textbf{Climate}: ``This location has a temperate oceanic climate."
    \item \textbf{Compass direction}: ``This photo is facing north."
    \item \textbf{Season (month)}: ``This photo was taken in December."
    \item \textbf{Traffic}: ``In this location, people drive on the left side of the road."
\end{itemize}
}\vspace{0.7em}

All the above caption components contain information relevant for the geolocalization of an image. Consequently, our continued contrastive pretraining creates an implicit multi-task setting and ensures the model learns rich representations of the data while learning features that are relevant to the task of image geolocalization.

\subsection{Multi-task learning with climate data}
We also experiment with making our multi-task setup explicit by creating task-specific prediction heads for auxiliary labels, and adapt our loss function according to \Cref{eq:loss_mt}, where $\mathcal{L}_{n, \text{loc}}$ corresponds to the loss in \Cref{eq:local_loss}. Our multi-task setup further includes a cross-entropy classification task ($\mathcal{L}_{n, \text{climate}}$) of the 28 different Köppen-Geiger climate zones \cite{beck_et_al_2018}, a cross-entropy month (season) classification task ($\mathcal{L}_{n, \text{month}}$), and six mean squared error (MSE) regression tasks (combined into $\mathcal{L}_{n, \text{reg}}$) that attempt to predict values related to the temperature, precipitation, elevation, and population density of a given location.

\begin{equation}
\normalsize
    \mathcal{L}_n = \mathcal{L}_{n, \text{loc}} + \alpha\mathcal{L}_{n, \text{climate}} + \beta\mathcal{L}_{n, \text{month}} + \gamma \mathcal{L}_{n, \text{reg}}
    \label{eq:loss_mt}
\end{equation}

%(1) the average temperate, (2) the average temperature difference between the hottest and coldest month, (3) the log elevation in meters, (4) the log precipitation (mm/day), (5) the log difference in precipitation (5) log population density (people/$km^2$), and (6)

%As climate variables we include the 28 Köppen-Geiger Climate Zone, the yearly average temperature and precipitation at the given location as well as the difference in temperature and precipitation between the month with the highest average value and the month with the lowest average value. The climate zone and and season prediction tasks are posed as a classification problem while the other six auxiliary tasks are formulated as a regression task.

We unfreeze the last CLIP layer to allow for parameter sharing across tasks with the goal of observing a positive transfer from our auxiliary tasks to our geolocalization problem and to learn more general image representations reducing the risk of overfitting to the training dataset. Adjusting $\alpha, \beta$, and $\gamma$, our loss function weighs the geolocalization task as much as all auxiliary tasks combined considering each task's loss magnitude. A novel contribution of our work is that we use a total of eight auxiliary prediction tasks instead of just two compared to prior research \cite{pramanick_et_al_2022}.

\subsection{Refinement via location cluster retrieval}
To further refine our model's guesses within a geocell and to improve street- and city-level performance, instead of simply predicting the mean latitude and longitude of all points within a geocell \cite{pramanick_et_al_2022}, we perform intra-geocell refinement. To this end, we design a hierarchical retrieval mechanism over location clusters akin to prototypical networks \cite{snell_2017} with fixed parameters. We again use the OPTICS clustering algorithm \cite{ankerst_1999} to cluster all points within a geocell $g$ and thus propose location clusters $C_g$ whose representation is the average of all corresponding image embeddings. To compute all image embeddings, we use our pretrained CLIP model $f(\cdot)$ described in \Cref{sec:pretraining}, mapping each image $l$ in a cluster $c$ to its embedding $f(l)$.

\begin{equation}
    c^* = \arg\min_{c \in C_g} \left\| f(x) - \frac{1}{|c|} \sum_{l \in c} f(l) \right\|_2 
    \label{eq:refine_1}
\end{equation}\\

During inference, we predict the location cluster $c^*$ of an input image $x$ by selecting the cluster with the minimum Euclidean image embedding distance to the input image embedding $f(x)$. Once the cluster $c^*$ is determined, we further refine our guess by choosing the single best location within the cluster, again via minimizing the Euclidean embedding distance. The retrieval over location clusters and within-cluster refinement add two additional levels of prediction hierarchy to our system, with the number of unique potential guesses equaling the training dataset size.

While hierarchical refinement via retrieval is in itself a novel idea, our work goes one step further. Instead of refining a geolocalization prediction within a single cell, our mechanism optimizes across multiple cells which further increases performance. During inference, our geocell classification model outputs the \textit{topK} predicted geocells (5 for PIGEON, 40 for PIGEOTTO) as well as the model's associated probabilities for these cells. The refinement model then picks the most likely location within each of the \textit{topK} proposed geocells, after which a softmax is computed across the \textit{topK} Euclidean image embedding distances. We use a temperature softmax with a temperature that is carefully calibrated on the validation datasets to balance probabilities across different geocells. Finally, these refinement probabilities are multiplied with the initial \textit{topK} geocell probabilities to determine a final location cluster and within-cluster refinement is performed as illustrated in~\Cref{fig:pipeline}.
\section{Experimental results and analysis}
\label{sec:experiments}

\subsection{Experimental setting}
\label{sec:experimental_setting}

\paragraph{Training PIGEON and PIGEOTTO.}

Based on our technical methodology outlined in~\Cref{sec:technical_approach}, we train two models for distinct downstream evaluation purposes.

First, inspired by GeoGuessr, we train PIGEON (Predicting Image Geolocations). We collect an original dataset of 100,000 randomly sampled locations from GeoGuessr and download a set of four images spanning an entire ``panorama" in a given location, or a 360-degree view, for a total of 400,000 training images. For each location, we start with a random compass direction and take four images separated by 90 degrees, carefully creating non-overlapping image patches.%Multiple images for a given geolocation oftentimes give us important additional information, greatly improving geolocalization performance.

%Recognizing that adding auxiliary geographic metadata can be beneficial for image geolocalization \cite{arbinger_et_al_2022}, we decided to augment our dataset with data on Köppen-Geiger climate zones \cite{beck_et_al_2018}, as well as elevation temperature, precipitation, and other data. We also capture information frequently used by human GeoGuessr players in placing their guesses such as the side of the road that traffic travels on.

% \textcolor{red}{Talk about the insight from PIGEON that made us train PIGEOTTO.}

Second, motivated by PIGEON's image geolocalization capabilities, we train PIGEOTTO (Predicting Image Geolocations with Omni-Terrain Training Optimizations). Unlike PIGEON, PIGEOTTO is not a Street View photo localizer but rather a general image geolocator. To that end, we access the MediaEval 2016 dataset~\cite{larson_et_al_2017} consisting of geo-tagged Flickr images from all over the world and obtain 4,166,186 images, considering that some images have become unavailable since 2016. Additionally, recognizing the importance of geolocating landmarks for general image geolocalization capabilities, we add 340,579 images from the Google Landmarks v2 dataset~\cite{weyand_et_al_2020} to our training mix which are all derived from Wikipedia. Importantly, there is no overlap in the training data we use between PIGEON and PIGEOTTO, as the models serve different downstream purposes. Unlike PIGEON, PIGEOTTO takes a single image per location as input, as obtaining a four-image panorama is often infeasible in general image geolocalization settings.
% (4.2M images from YFCC + ~200k images from Wikipedia (this is the google landmarks v2 dataset))
% TODO: How was MP-16 filtered down?
% TODO: How were landmarks chosen?
% Evolved into PIGEOTTO.

\paragraph{Evaluation datasets and metrics.}

Our work defines the median distance error to the correct location as the primary and composite metric. In line with the prior literature on image geolocalization, we further evaluate the ``\% @ km" statistic in our analysis as a more fine-grained metric. For a given dataset, the ``\% @ km" statistic determines the percentage of guesses that fall within a given kilometer-based distance from the ground-truth location. Just as in the prior work, we evaluate five distance radii: 1 km (roughly street-level accuracy), 25 km (city-level), 200 km (region-level), 750 km (country-level), and 2,500 km (continent-level).

For PIGEON, we run evaluations on a holdout dataset collected from GeoGuessr consisting of 5,000 Street View locations. We separately conduct extensive blind experiments in GeoGuessr deploying PIGEON against human players with varying degrees of expertise as well as a separate match against a world-class professional player. To quantify which parts of our modeling setup impact performance, we further run eight separate ablation studies. %to quantify the impact of our technical contributions.

For PIGEOTTO, we focus our evaluations squarely on the benchmark datasets that are established in the literature. Namely, we look at IM2GPS~\cite{hays_and_efros_2008}, IM2GPS3k~\cite{vo_et_al_2017}, YFCC4k~\cite{vo_et_al_2017} and YFCC26k~\cite{mueller_budack_et_al_2018} (based on the MediaEval 2016 dataset~\cite{larson_et_al_2017}), and GWS15k~\cite{clark_et_al_2023}. As the last dataset has not been publicly released by the time of this writing, we reconstruct the dataset by exactly replicating the dataset generation procedure outlined in \citet{clark_et_al_2023}.

%-------------------------------------------------------------------------
\subsection{Street View evaluation with PIGEON}
\label{sec:pigeon_results}
We present the results of our evaluations of PIGEON and ablations of our contributions in~\Cref{table:ablation_study} and~\Cref{table:additional_results_distance}. As evidenced by our results, each subsequent ablation deteriorates most metrics, pointing to the synergistic nature of the ensemble of methods in our geolocalization system.

%-------------------------------------------------------------------------
% Table: PIGEON Ablations
%-------------------------------------------------------------------------
\begin{table}[!htpb]
\centering
\caption{Cumulative ablation study of our image geolocalization system on a holdout dataset of 5,000 Street View locations.}
\vspace{10pt}
\begin{tabular}{lcccc}
\toprule
                                           &  \textbf{Country} &  \textbf{Mean}     &  \textbf{Median}  & \textbf{Geoguessr}                \\
 \textbf{Ablation}                           & \textbf{Accuracy} & \textbf{Error}  & \textbf{Error} & \textbf{Score}\\
                                           & $\%$       & $km$        & $km$       & $points$\\
\midrule
    \textbf{PIGEON}  & \textbf{91.96}  & \textbf{251.6} & \textbf{44.35}  & 4,525\\
    \midrule
    $-$ Freezing Last CLIP Layer After Pretraining   & 91.82       & 255.1      & 45.47       & \textbf{4,531}\\
    $-$ Hierarchical Guess Refinement       & 91.14       & 251.9      & 50.01       & 4,522      \\
    $-$ Contrastive CLIP Pretraining                  & 89.36       & 316.9      & 55.51       & 4,464    \\
    $-$ Semantic Geocells       & 87.96       & 299.9      & 60.63       & 4,454         \\
    $-$ Multi-task Prediction Heads        & 87.90       & 312.7      & 61.81       & 4,442          \\
    $-$ Fine-tuning Last CLIP Layer  & 87.64   & 315.7  & 60.81       & 4,442          \\
    $-$ Four-image Panorama  & 74.74       & 877.4      & 131.1       & 3,986          \\
    $-$ Haversine Smoothing    & 72.12       & 990.0      & 148.0       & 3,890\\
      
\bottomrule
\end{tabular}
\label{table:ablation_study}
\end{table}

%-------------------------------------------------------------------------
% Table: PIGEON Ablations Beyond Distance
%-------------------------------------------------------------------------
\begin{table}[!htbp]
\centering
\caption{Cumulative ablation study using five common distance radii on a holdout dataset of 5,000 Street View locations.}
\vspace{10pt}
\begin{tabular}{lccccc}
\toprule
   &  \multicolumn{5}{c}{\textbf{Distance (\% @ km)}}                                    \\
 \textbf{Ablation}      & \textit{Street} & \textit{City}  & \textit{Region} & \textit{Country} & \textit{Continent}  \\
            & 1 km   & 25 km & 200 km & 750 km  & 2,500 km    \\
\midrule
    \textbf{PIGEON} &    \textbf{5.36}    &   \textbf{40.36}    &    78.28    &    94.52     &      \textbf{98.56}    \\
    \midrule
    $-$ Freezing Last CLIP Layer After Pretraining &    4.84    &   39.86    &    \textbf{78.98}    &    94.76     &      98.48    \\
    $-$ Hierarchical Guess Refinement  &    1.32    &   34.96    &    78.48    &    \textbf{94.82}     &      98.48     \\ % without fine-tuning last layer
    $-$ Contrastive CLIP Pretraining   &    1.24    &   34.54    &    76.36    &    93.36     &      97.94     \\
    $-$ Semantic Geocells    &    1.18    &   33.22    &    75.42    &    93.42     &      98.16     \\
    $-$ Multi-task Prediction Heads  &    1.10    &   32.74    &    75.14    &    93.00     &      97.98      \\
    $-$ Fine-tuning Last CLIP Layer   &    1.10    &   32.50    &    75.32    &    92.92     &      98.00      \\
    $-$ Four-image Panorama &    0.92    &   24.18    &    59.04    &    82.84     &      92.76      \\
    $-$ Haversine Smoothing &    1.28    &   24.08    &    55.38    &    80.20     &      92.00\\

\bottomrule
\end{tabular}
\label{table:additional_results_distance}
\end{table}

Starting from the very bottom of both tables, corresponding to a simple CLIP vision encoder plus a geocell prediction head, we can see that with the introduction of haversine smoothing, the mean distance error decreases by 112.6 kilometers from 990.0 to 877.4 kilometers. The bulkiest performance lift, however, comes from the introduction of a four-image panorama instead of a single image, increasing our country accuracy by 12.9 percentage points and more than halving our median kilometer error from 131.1 to 60.8 kilometers. While fine-tuning the last CLIP layer and sharing parameters in a multi-task setting slightly improves the performance of our model, the uplift is much more palpable with the introduction of our semantic geocells, reducing the median error from 60.6 to 55.5 kilometers. When we additionally pretrain CLIP via our synthetic captions, we gain another 1.7 percentage points in long-range country accuracy. Complemented by our hierarchical location cluster refinement, we improve short-range street-level accuracy from 1.3\% to 4.8\%. Finally, we freeze the last CLIP layer again and thus prevent parameter sharing between our geocell and multi-task prediction heads, given that our pretraining procedure already incorporates multi-task training. This results in PIGEON's final metrics of a 92.0\% country accuracy and a median distance error of 44.4 kilometers.

%-------------------------------------------------------------------------
% Figure: Geoguessr Players
%-------------------------------------------------------------------------
\begin{figure}[!htbp]
\includegraphics[width=\textwidth]{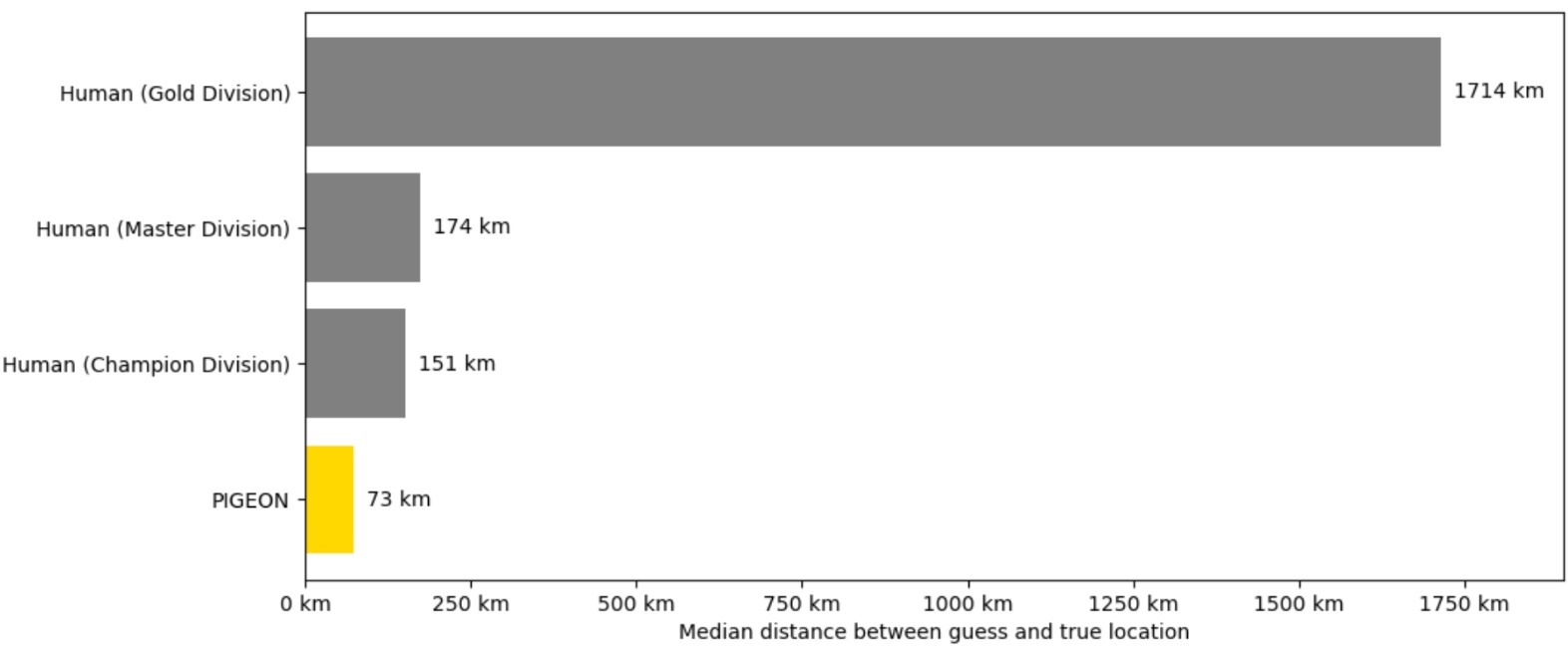}
\caption{Geolocalization error of PIGEON against human players of various in-game skill levels across 458 multi-round matches. The Champion Division consists of the top 0.01\% of players. PIGEON's error is higher than in \Cref{table:ablation_study} because GeoGuessr round difficulties are adjusted dynamically, increasing with every round.}
\label{fig:geoguessr_players}
\end{figure}

Beyond our ablations, we compare PIGEON's performance to humans in the game of GeoGuessr. To do so, we develop a Chrome extension bot that has access to PIGEON as an API and deploy our system in a blind experiment across 458 matches, each consisting of multiple rounds. PIGEON comfortably outperforms players in GeoGuessr's Champion Division, consisting of the top 0.01\% of human players. The results are shown in~\Cref{fig:geoguessr_players}, underscoring PIGEON's ability to beat players of all skill levels. Notably, top GeoGuessr players perform orders of magnitudes better than the players evaluated in \citet{seo_et_al_2018}.

For our final evaluation, we challenge one of the world's foremost professional GeoGuessr players to a match and win six out of six planet-scale, multi-round games.\footnote{\url{https://www.youtube.com/watch?v=ts5lPDV--cU}.} PIGEON is the first model to reliably beat a GeoGuessr professional.

%-------------------------------------------------------------------------
\subsection{Benchmark evaluation with PIGEOTTO}
\label{sec:pigeon_results}

%-------------------------------------------------------------------------
% Table: PIGEOTTO Results
%-------------------------------------------------------------------------
\begin{table}
    \centering
    \caption{Comparison of PIGEOTTO's results against other models on benchmark datasets. PIGEOTTO reduces the median kilometer error by 2-5x on benchmarks not solely focused on landmarks.}
    \vspace{10pt}
    \resizebox{1.0\textwidth}{!}{
    \begin{tabular}{c|c|c|ccccc}
        \toprule
        \multirow{3}{*}{\textbf{Benchmark}} & \multirow{3}{*}{\textbf{Method}} & \textbf{Median} & 
        \multicolumn{5}{c}{\textbf{Distance (\% @ km)}} \\
        % \multicolumn{5}{c}{\textbf{Distance} $(\alpha_r \textbf{[\%] @ km})$} & \textbf{Median} \\
        & & \textbf{Error} & \textit{Street} & \textit{City} & \textit{Region} & \textit{Country} & \textit{Continent}\\
        & & km & \textbf{1 km} & \textbf{25 km} & \textbf{200 km} & \textbf{750 km} & \textbf{2,500 km} \\
        
        \midrule
        
        \multirow{7}{*}{\textbf{IM2GPS} \cite{hays_and_efros_2008}} 
        & PlaNet \cite{weyand_et_al_2016} & $> 200$ & 8.4 & 24.5 & 37.6 & 53.6 & 71.3\\
        & CPlaNet \cite{seo_et_al_2018} & $> 200$ & 16.5 & 37.1 & 46.4 & 62.0 & 78.5\\
        %& ISNs(M, $f, S_3$) [11] & 16.5 & 42.2 & 51.9 & 66.2 & 81.0 \\
        & ISNs(M,$f^*$,$S_3$) \cite{mueller_budack_et_al_2018} & $> 25$ & 16.9 & 43.0 & 51.9 & 66.7 & 80.2 \\
        & Translocator \cite{pramanick_et_al_2022} & $> 25$ & 19.9 & 48.1 & 64.6 & 75.6 & 86.7\\
        & GeoDecoder \cite{clark_et_al_2023} & \textbf{$\sim$ 25} & \textbf{22.1} & \textbf{50.2} & \textbf{69.0} & 80.0 & 89.1\\
        & \textbf{PIGEOTTO (Ours)} & 70.5 & 14.8 & 40.9 & 63.3 & \textbf{82.3} & \textbf{91.1}\\
        \cline{2-8}
        & $\Delta$ {\scriptsize (\%} points{\scriptsize)} & & -7.3 & -9.3 & -5.7 & +2.3 & +2.0 \\
        
        \midrule
        
        \multirow{7}{*}{\textbf{IM2GPS3k} \cite{vo_et_al_2017}} & PlaNet \cite{weyand_et_al_2016} & $> 750$ & 8.5 & 24.8 & 34.3 & 48.4 & 64.6\\
        & CPlaNet \cite{seo_et_al_2018} & $> 750$ & 10.2 & 26.5 & 34.6 & 48.6 & 64.6\\
        %& ISNs(M, $f, S_3$) [11] & 10.1 & 27.2 & 36.2 & 49.3 & 65.6 \\
        & ISNs(M,$f^*$,$S_3$) \cite{mueller_budack_et_al_2018} & $\sim 750$ & 10.5 & 28.0 & 36.6 & 49.7 & 66.0 \\
        & Translocator \cite{pramanick_et_al_2022} & $> 200$ & 11.8 & 31.1 & 46.7 & 58.9 & 80.1\\
        & GeoDecoder \cite{clark_et_al_2023} & $> 200$ & \textbf{12.8} & 33.5 & 45.9 & 61.0 & 76.1 \\
        & \textbf{PIGEOTTO (Ours)} & \textbf{147.3} & 11.3 & \textbf{36.7} & \textbf{53.8} & \textbf{72.4} & \textbf{85.3} \\
        \cline{2-8}
        & $\Delta$ {\scriptsize (\%} points{\scriptsize)} & & -1.5 & +3.2 & +7.9 & +11.4 & +9.2 \\
        
        \midrule
        
        \multirow{7}{*}{\textbf{YFCC4k} \cite{vo_et_al_2017}} & PlaNet \cite{weyand_et_al_2016} & $> 750$ & 5.6 & 14.3 & 22.2 & 36.4 & 55.8 \\
        & CPlaNet \cite{seo_et_al_2018} & $> 750$ & 7.9 & 14.8 & 21.9 & 36.4 & 55.5  \\
        %& ISNs(M, $f, S_3$) [11] & 6.5 & 16.2 & 23.8 & 37.4 & 55.0 \\
        & ISNs(M,$f^*$,$S_3$) \cite{mueller_budack_et_al_2018} & $> 750$ & 6.7 & 16.5 & 24.2 & 37.5 & 54.9\\
        & Translocator \cite{pramanick_et_al_2022} & $> 750$ & 8.4 & 18.6 & 27.0 & 41.1 & 60.4\\
        & GeoDecoder \cite{clark_et_al_2023} & $\sim 750$ & 10.3 & \textbf{24.4} & 33.9 & 50.0 & 68.7 \\
        & \textbf{PIGEOTTO (Ours)} & \textbf{383.0} & \textbf{10.4} & 23.7 & \textbf{40.6} & \textbf{62.2} & \textbf{77.7}\\
        \cline{2-8}
        & $\Delta$ {\scriptsize (\%} points{\scriptsize)} & & +0.1 & -0.7 & +6.7 & +12.2 & +9.0 \\

        \midrule

        \multirow{6}{*}{\textbf{YFCC26k} \cite{mueller_budack_et_al_2018}} & PlaNet \cite{weyand_et_al_2016} & $> $ 2,500 & 4.4 & 11.0 & 16.9 & 28.5 & 47.7\\
        %& ISNs(M, $f, S_3$) [11] & 5.3 & 12.1 & 18.8 & 31.8 & 50.6 \\
        & ISNs(M,$f^*$,$S_3$) \cite{mueller_budack_et_al_2018} & $\sim$ 2,500 & 5.3 & 12.3 & 19.0 & 31.9 & 50.7 \\
        & Translocator \cite{pramanick_et_al_2022} & $>750$ & 7.2 & 17.8 & 28.0 & 41.3 & 60.6\\
        & GeoDecoder \cite{clark_et_al_2023} & $\sim 750$ & 10.1 & 23.9 & 34.1 & 49.6 & 69.0\\
        & \textbf{PIGEOTTO (Ours)} & \textbf{333.3} & \textbf{10.5} & \textbf{25.8} & \textbf{42.7} & \textbf{63.2} & \textbf{79.0}\\
        \cline{2-8}
        & $\Delta$ {\scriptsize (\%} points{\scriptsize)} & & +0.4 & +1.9 & +8.6 & +13.6 & +10.0 \\
        
        \midrule

        \multirow{5}{*}{\textbf{GWS15k} \cite{clark_et_al_2023}}
        & ISNs(M,$f^*$,$S_3$) \cite{mueller_budack_et_al_2018} & $>$ 2,500 & 0.05 & 0.6 & 4.2 & 15.5 & 38.5\\
        & Translocator \cite{pramanick_et_al_2022} & $>$ 2,500 & 0.5 & 1.1 & 8.0 & 25.5 & 48.3\\
        & GeoDecoder \cite{clark_et_al_2023} & $\sim$ 2,500 & \textbf{0.7} & 1.5 & 8.7 & 26.9 & 50.5 \\
        & \textbf{PIGEOTTO (Ours)} & \textbf{415.4} & \textbf{0.7} & \textbf{9.2} & \textbf{31.2} & \textbf{65.7} & \textbf{85.1}\\
        \cline{2-8}
        & $\Delta$ {\scriptsize (\%} points{\scriptsize)} & & +0.0 & +7.7 & +22.5 & +38.8 & +34.6 \\
        \bottomrule
    \end{tabular}
    }
    \label{table:pigeotto_results}
\end{table}

The results of our evaluations of PIGEOTTO on benchmark datasets are displayed in \Cref{table:pigeotto_results}. PIGEOTTO achieves state-of-the-art (SOTA) performance on every single benchmark dataset and on the majority of distance-based granularities. On IM2GPS, it is able to improve the state of the art on both country-level and continent-level accuracy by 2 percentage points or more. Its relative underperformance on smaller granularities can be attributed to the landmark-only nature of IM2GPS and its small size of 237 images. On a larger and more general dataset, IM2GPS3k, PIGEOTTO performs much better, achieving SOTA performance on all but the street-level metric, with an impressive 11.4 percentage-point improvement on the country level and a much lower median error of 147.3 kilometers. Meanwhile, on YFCC4k and YFCC26k, PIGEOTTO is able to outperform the current state of the art on 9 out of 10 metrics, including by 12.2 percentage points on the country level on YFCC4k and by 13.6 percentage points on YFCC26k, more than halving the previous SOTA median error. Finally, we see very significant improvements on the most recently released benchmark, GWS15k, consisting entirely of Street View images. Crucially, GWS15k is the most difficult dataset in the benchmark set. If we define images to be taken in the same location if they are less than 100 meters apart, 92\% of locations in GWS15k are not taken in the same location as any MediaEval 2016~\cite{larson_et_al_2017} training data on which prior SOTA models and our system were trained. For comparison, this number ranges from 23\% to 42\% for the other four benchmark datasets, underscoring the unique difficulty of GWS15k. Noting that PIGEOTTO was not trained on any Street View images, this suggests that PIGEOTTO is truly planet-scale in nature, exhibits robust behavior to distribution shifts, and is the first geolocalization model that effectively generalizes to unseen places.

\section{Ethical considerations}
\label{sec:ethics}

Image geolocalization represents a sub-discipline of computer vision that comes with both potential benefits to society as well as with risks of misuse. While prior work in the field addresses ethical implications scantily, we believe that the potential misuse and negative downstream implications of image geolocalization systems afford a separate discussion section in this paper.

On the one hand, accurate geo-tagging of images opens up possibilities for various beneficial applications, far beyond the game of GeoGuessr, including helping to understand changes to particular locations over time. Image geolocalization has found use cases in autonomous driving, navigation, geography education, open-source intelligence, and visual investigations in journalism.

On the other hand, however, applications of image geolocalization may come with risks, especially if the precision of such systems significantly improves in the future. To our knowledge, this is the first state-of-the-art image geolocalization paper in the last five years that is not funded by military contracts. Recently published work has been supported by grants from the Department of Defense~\cite{pramanick_et_al_2022} and the US Army~\cite{clark_et_al_2023}. Any attempts to develop image geolocalization technology for military use cases should come under particular scrutiny. There are also privacy risks involved; for instance, some methods using Street View images have been shown to be capable of inferring local income, race, education, and voting patterns \cite{gebru_et_al_2017}.

Image geolocalization technologies come with dual-use risks~\cite{henderson_et_al_2023}, and efforts need to be made to minimize harmful consequences. To that end, we decide not to release model weights publicly and only release our code for academic validation. While a major limitation of today's image geolocalization technologies (including ours) is that they are unable to make street-level predictions reliably, researchers ought to carefully consider the risk of potential misuse of their work as such technologies get increasingly precise.

\section{Conclusion}
\label{sec:conclusion}

We propose a novel deep multi-task approach for planet-scale image geolocalization that achieves state-of-the-art benchmark results while being robust to distribution shifts.

To confirm the efficacy of our approach, we train and evaluate two distinct image geolocalization models. First, we gather a global Street View dataset to train PIGEON, a multi-task model that places into the top 0.01\% of human players in the game of GeoGuessr. On a holdout dataset of 5,000 Street View locations, 40.4\% of PIGEON's predictions of geographic coordinates land within a 25-kilometer radius of the ground-truth location. Subsequently, we assemble a planet-scale dataset of over 4 million images derived from Flickr and Wikipedia to train the more general PIGEOTTO, improving the state of the art on a wide range of geolocalization benchmark datasets by a large margin.

Going forward, it remains to be seen whether applied image geolocalization technologies will be truly planet-scale or focused on a well-defined narrow distribution. In any case, our findings about the importance of semantic geocell creation, multimodal contrastive pretraining, and precise intra-geocell refinement, among others, point to important building blocks for such systems. Nevertheless, deployment of any downstream image geolocalization technology will need to balance potential benefits with possible risks, ensuring the responsible development of future computer vision systems.

\bibliography{paper}  %%% Uncomment this line and comment out the ``thebibliography'' section below to use the external .bib file (using bibtex) .





% \end{thebibliography}

\newpage
\appendix

\section*{Appendix}
\label{sec:supplementary_material_contents}

We include additional details in this appendix. Specifically, we expand on the following topics:\\

A. Semantic geocell creation

B. Implementation details

C. Ablation study on pretraining captions

D. Ablation study on training datasets

E. Auxiliary data sources

F. Ablation studies on non-distance metrics

G. Additional analyses

H. Deployment to Geoguessr

I. Acknowledgements

%-------------------------------------------------------------------------

\section{Semantic geocell creation}
\label{sec:semantic_geocell_creation}

In the body of our work, we described how our semantic geocell creation algorithm works on a high level. Similar to approaches in prior literature such as \citet{theiner_et_al_2021}, we create a hierarchy of administrative areas and merge adjacent geocells until a set minimum number of training samples per geocell is reached. This, however, results in a highly imbalanced classification problem, especially for larger training datasets. A major contribution of our work is that we define a method to split larger geocells into smaller, still semantically meaningful cells, by leveraging the information contained in the training data's geolocations. The key insight is that locations from most training distributions tend to cluster around popular places and landmarks, and these clusters can be extracted.\\

\Cref{alg:semantic_geocell_creation} shows a slightly simplified version of how we split large geocells into multiple smaller ones without the help of administrative boundary information, resulting in a much more balanced geocell classification dataset. As one can see, the algorithm only depends on the geocell boundaries or shape definitions $g$, the training dataset $x$, an OPTICS clustering algorithm with parameters $p$ (optionally round-specific parameters $p_j$), and a minimum cell size MINSIZE. The VORONOI algorithm takes a set of points as input and outputs a new geocell shape defined by these points which can be removed from the original cell shape.\\

\begin{figure}[!htbp]
\centering
\includegraphics[width=0.7\textwidth]{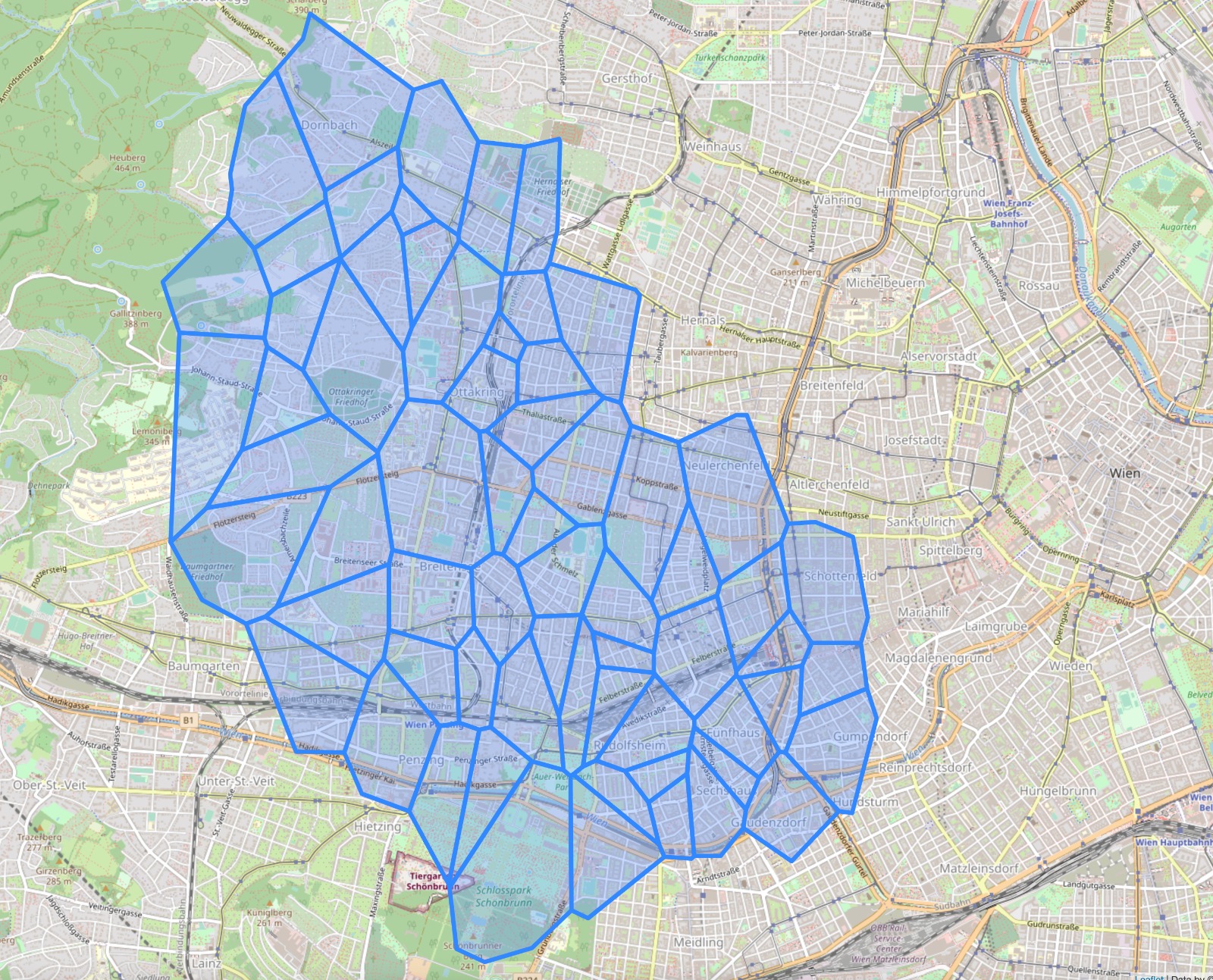}
\vspace{5pt}
\caption{Voronoi tessellation applied in the process of geocell creation for points of an OPTICS cluster in Vienna, Austria, based on political boundaries from GADM \cite{gadm_2022}.}
\label{fig:voronoi_tessellation}
\end{figure}

\Cref{fig:voronoi_tessellation} shows a small geocell that has been extracted from a larger geocell covering the entire city of Vienna, Austria, via Voronoi tessellation. The partitions within the blue geocells are the result of the Voronoi tesselation algorithm assigning to each training sample all geographic area to which it is closest. 

\begin{algorithm}[!htbp]
   \caption{Simplified Semantic Geocell Splitting}
   \label{alg:semantic_geocell_creation}
\begin{algorithmic}
   \STATE {\bfseries Input:} geocell boundaries $g$, training samples $x$,\\
    OPTICS parameters $p$, minimum cell size MINSIZE.
   \STATE Initialize $j = 1$.
   \REPEAT
   \STATE Initialize $C$ = OPTICS($p_j$).
   \FOR{$g_i$ {\bfseries in} $g$}
   \STATE Define $x_i = \{x_k | x_k \in x \land x_k \in g_i\}$.
   \REPEAT
   \STATE Cluster $c = C(x_i)$.
   \STATE $c_{max} = c_{k}$ where $|x_{i, k}| \geq |x_{i, l}| \forall l$.
   \IF{$|c_{max}| > $ MINSIZE and $|x \setminus x_{i, k}| > $ MINSIZE}
   \STATE New cell $g_{new}$ = VORONOI($x_{i, k}$).
   \STATE $g_{i} = g_{i} \setminus g_{new}$.
   \STATE Assign $x_i$ to cells $i$ and $new$, respectively.
   \ENDIF
   \UNTIL{convergence}\\
   \ENDFOR\\
   \STATE $j = j + 1$
   \UNTIL{$j$ is $|p|$}
\end{algorithmic}
\end{algorithm}

\Cref{fig:voronoi_tessellation} shows a small geocell that has been extracted from a larger geocell covering the entire city of Vienna, Austria, via Voronoi tessellation. The partitions within the blue geocells are the result of the Voronoi tesselation algorithm assigning to each training sample all geographic area to which it is closest. 

\section{Implementation details}
In this section, we describe the implementation details of PIGEON and PIGEOTTO and further illustrate how the two models differ from each other.

\subsection{Model input}

\begin{figure}[!htbp]
\centering
\includegraphics[width=1.0\textwidth]{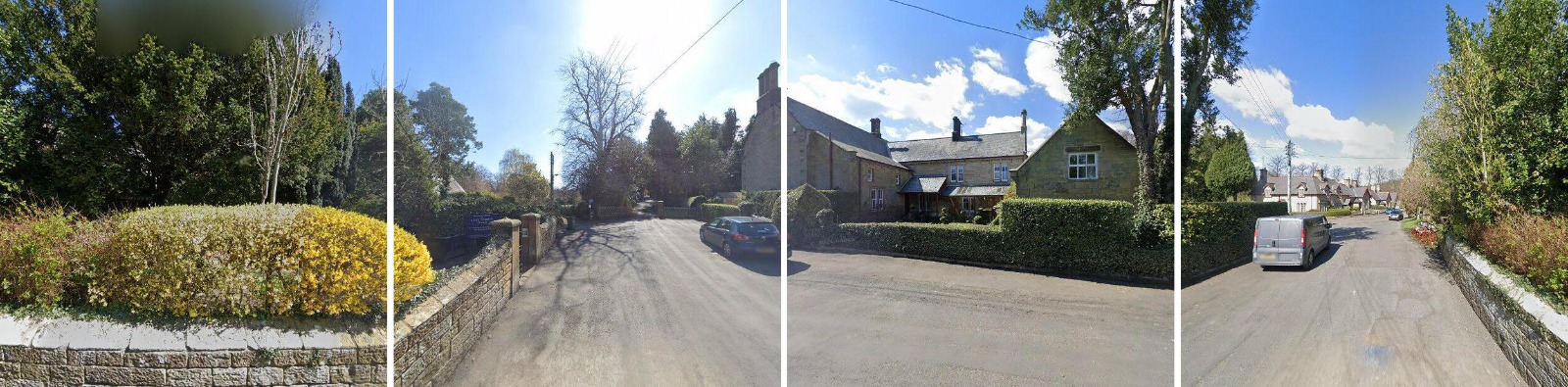}
\caption{Four images comprising a 360-degree panorama from a location in Pegswood, England, in our dataset.}
\label{fig:panorama}
\end{figure}

The biggest difference between PIGEON and PIGEOTTO is that PIGEON takes a four-image Street View panorama as input, whereas PIGEOTTO takes a single image as input. Images are always cropped to a square aspect ratio before being fed into the models. \Cref{fig:panorama} shows a representative input for PIGEON, depicting a 360-degree, four-image Street View panorama taken in Pegswood, England.

\begin{figure}[!htbp]
\centering
\includegraphics[width=0.9\textwidth]{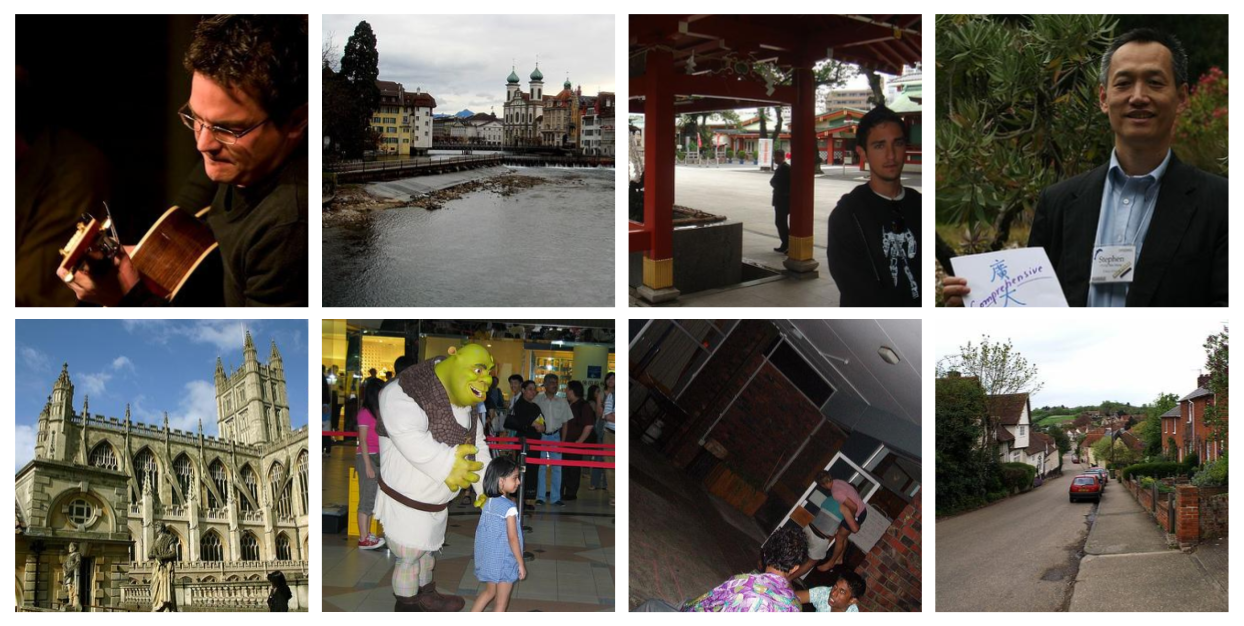}
\caption{Eight samples from the MediaEval 2016 dataset \cite{larson_et_al_2017}.}
\label{fig:media_eval_sample_images}
\end{figure}

PIGEOTTO's training dataset is vastly different to PIGEON's Street View input; the model takes a single image as input and was trained on a highly diverse image geolocalization dataset. \Cref{fig:media_eval_sample_images} shows eight images sampled from the MediaEval 2016 dataset \cite{larson_et_al_2017} which was derived from user-uploaded Flickr images. It is clearly apparent that some of the images are extremely difficult to geolocalize, for example because they were taken indoors.

\subsection{Pretraining}
\Cref{table:trainargs_pretraining} shows the hyperparameter settings employed for our contrastive pretraining of CLIP for the task of image geolocalization. The CLIP weights were initialized with the pretrained weights of OpenAI's CLIP implementation.\footnote{\url{https://huggingface.co/openai/clip-vit-large-patch14-336}.}

\begin{table}[!htpb]
\centering
\caption{Hyperparameter settings for pretraining CLIP's vision encoder for the task of image geolocalization.}
\vspace{10pt}
\label{table:trainargs_pretraining}
\begin{tabular}{lcc}
\toprule
\textbf{Parameter} & \textbf{PIGEON} & \textbf{PIGEOTTO} \\
\midrule
GPU Type & A100 80GB & A100 80GB\\
Number of GPUs & 4 & 4\\
Dataset Source & Street View & Flickr\\
Dataset Size (Samples) & $\sim$ 1M & $\sim$ 4.2M\\
Batch Size & 32 & 32\\
Gradient Accumulation Steps & 8 & 8\\
Optimizer & AdamW & AdamW\\
Learning Rate & $1e^{-6}$ & $5e^{-7}$\\
Weight Decay & $1e^{-3}$ & $1e^{-3}$\\
Warmup (Epochs) & 0.2 & 0.02\\
Training Epochs & 3 & 2\\
Adam $\beta_1$ & 0.9 & 0.9\\
Adam $\beta_2$ & 0.98 & 0.98\\
\bottomrule
\end{tabular}
\end{table}

\subsection{Fine-tuning}

The fine-tuning of PIGEON and PIGEOTTO consists of adding a linear layer on top of the pretrained vision encoder, mapping image embeddings to a fixed number of geocells. During this process, the weights of the vision encoder remain frozen. \Cref{table:trainargs_finetuning} shows the hyperparameters used in this training step. Both PIGEON and PIGEOTTO were trained until convergence.

\begin{table}[!htpb]
\centering
\caption{Hyperparameter settings for fine-tuning CLIP's vision encoder via a linear projection layer onto geocells.}
\vspace{10pt}
\label{table:trainargs_finetuning}
\begin{tabular}{lcc}
\toprule
\textbf{Parameter} & \textbf{PIGEON} & \textbf{PIGEOTTO} \\
\midrule
GPU Type & A100 80GB & A100 80GB\\
Number of GPUs & 1 & 1\\
Dataset Source & Street View & Flickr + Wikipedia\\
Dataset Size (Samples) & $\sim$ 100k & $\sim$ 4.5M\\
Number of Geocells & 2,203 & 2,076\\
Haversine Smoothing $\tau$ & 75 & 65\\
Batch Size & 1024 & 1024\\
Gradient Accumulation Steps & 1 & 1\\
Optimizer & AdamW & AdamW\\
Learning Rate & $5e^{-5}$ & $2e^{-5}$\\
Weight Decay & 0.01 & 0.01\\
Training Epochs & Convergence & Convergence\\
Adam $\beta_1$ & 0.9 & 0.9\\
Adam $\beta_2$ & 0.999 & 0.999\\
\bottomrule
\end{tabular}
\end{table}

\pagebreak

\subsection{Hierarchical refinement}

We use a hierarchical retrieval mechanism over location clusters to refine predictions. As a first step, location clusters are pre-computed using an OPTICS clustering algorithm. Then, during inference, a cluster is selected according to \Cref{eq:refine_1}. Finally, the location guess is refined within the top selected cluster. The refinement process is also dependent on a number of parameters, the most important of which are listed in \Cref{table:trainargs_refinement} and contrasted between PIGEON and PIGEOTTO.

\begin{table}[!htpb]
\centering
\caption{Parameters used in our hierarchical retrieval mechanism over location clusters.}
\vspace{10pt}
\label{table:trainargs_refinement}
\begin{tabular}{lcc}
\toprule
\textbf{Parameter} & \textbf{PIGEON} & \textbf{PIGEOTTO} \\
\midrule
Number of Geocell Candidates & 5 & 40\\
Maximum Refinement Distance (km) & 1,000 & None\\
Distance Metric & Euclidian & Euclidian\\
Softmax Temperature & 1.6 & 0.6\\
OPTICS Min Samples (Cluster Creation) & 3 & 10\\
OPTICS xi (Cluster Creation) & 0.15 & 0.1\\
\bottomrule
\end{tabular}
\end{table}

%-------------------------------------------------------------------------
\section{Ablation study on pretraining captions}

In \Cref{sec:pretraining}, we describe a novel multi-task contrastive pretraining method for image geolocalization. The ablation in \Cref{table:caption_ablation} shows that our pretraining reduces PIGEON's median kilometer error significantly from 57.8 to 44.4 kilometers ($-23.3\%$) versus no pretraining as in \citet{wu_and_huang_2022}. Our innovation is that we are the first to design a multi-modal \textit{and} multi-task contrastive pretraining objective for CLIP through the use of synthetic captions, and further find that the multi-task component of our method is highly effective; we observe a positive transfer from the auxiliary tasks embedded in our captions to the task of geolocalization, reducing our median error from 49.4 to 44.4 kilometers ($-10.2\%$) compared to pretraining solely with location captions as in \citet{haas_et_al_2023}. Our multi-task contrastive pretraining method is general enough that it could also be employed in other problem domains.

\begin{table}[htp]
    \centering
    \caption{Ablation study of CLIP pretraining captions for PIGEON on a holdout dataset of 5,000 Street View locations.}
    \vspace{10pt}
    \resizebox{1.0\textwidth}{!}{
    \begin{tabular}{l|c|ccccc}
        \toprule
        \toprule
        \multirow{2}{*}{\textbf{Ablation}} & \textbf{Median Error} & 
        \multicolumn{5}{c}{\textbf{Distance (\% @ km)}} \\
         & km & \textbf{1 km} & \textbf{25 km} & \textbf{200 km} & \textbf{750 km} & \textbf{2,500 km} \\
        
        \midrule
        PIGEON \textbf{[location + auxiliary captions]} & \textbf{44.35} & \textbf{5.36} & \textbf{40.36} & \textbf{78.28} & \textbf{94.52} & \textbf{98.56}\\
        \midrule
        PIGEON \textbf{[location captions} as in \cite{haas_et_al_2023}\textbf{]} & 49.37 & 4.62 & 38.46 & 77.10 & 94.34 & 98.48\\
        PIGEON \textbf{[no pretraining} as in \cite{wu_and_huang_2022}\textbf{]} & 57.80 & 4.48 & 36.18 & 74.88 & 93.24 & 98.04\\
        \bottomrule
        \bottomrule
    \end{tabular}
    }
    \label{table:caption_ablation}
\end{table}

%-------------------------------------------------------------------------

\section{Ablation study on training datasets}
\Cref{sec:experimental_setting} in the body of our paper describes the different datasets used to train PIGEON and PIGEOTTO. While PIGEON was purely trained on Street View imagery, the training dataset for PIGEOTTO contains a combination of 4,166,186 geo-tagged images from the MediaEval 2016 dataset~\cite{larson_et_al_2017} and 340,579 images from the Google Landmarks v2 dataset~\cite{weyand_et_al_2020}. Prior works' benchmark results, listed in \Cref{table:pigeotto_results}, employ a diverse range of training datasets with the goal of building the best performing and robust image geolocalization models. Since the prior SOTA model Geodecoder~\cite{clark_et_al_2023} was exclusively trained on the MediaEval 2016 dataset~\cite{larson_et_al_2017}, we include an additional training dataset ablation for PIGEOTTO in \Cref{table:no_landmarks} to distinguish data selection from system design effects.

\begin{table}[htp]
    \centering
    \caption{Ablation study of PIGEOTTO's Google Landmarks v2 ~\cite{weyand_et_al_2020} data (340k images) against prior SOTA on five benchmarks.}
    \vspace{10pt}
    \resizebox{1.0\textwidth}{!}{
    \begin{tabular}{l|l|c|ccccc}
        \toprule
        \toprule
        \multirow{3}{*}{\textbf{Benchmark}} & \multirow{3}{*}{\textbf{Method}} & \textbf{Median} & 
        \multicolumn{5}{c}{\textbf{Distance (\% @ km)}} \\
        & & \textbf{Error} & \textit{Street} & \textit{City} & \textit{Region} & \textit{Country} & \textit{Continent}\\
        & & km & \textbf{1 km} & \textbf{25 km} & \textbf{200 km} & \textbf{750 km} & \textbf{2,500 km} \\
        
        \midrule
        
        \multirow{3}{*}{\textbf{IM2GPS} \cite{hays_and_efros_2008}} 
        & GeoDecoder \cite{clark_et_al_2023} & \textbf{$\sim$ 25} & \textbf{22.1} & \textbf{50.2} & \textbf{69.0} & 80.0 & 89.1\\
        & PIGEOTTO \textbf{[ME16]} & 75.6 & 11.8 & 38.8 & 63.7 & 80.6 & \textbf{91.1}\\
        & PIGEOTTO \textbf{[ME16 + Landmarks]} & 70.5 & 14.8 & 40.9 & 63.3 & \textbf{82.3} & \textbf{91.1}\\
        
        \midrule
        
        \multirow{3}{*}{\textbf{IM2GPS3k} \cite{vo_et_al_2017}} 
        & GeoDecoder \cite{clark_et_al_2023} & $> 200$ & \textbf{12.8} & 33.5 & 45.9 & 61.0 & 76.1 \\
        & PIGEOTTO \textbf{[ME16]} & 163.6 & 10.9 & 35.8 & 52.4 & 70.7 & 84.4\\
        & PIGEOTTO \textbf{[ME16 + Landmarks]} & \textbf{147.3} & 11.3 & \textbf{36.7} & \textbf{53.8} & \textbf{72.4} & \textbf{85.3} \\
        
        \midrule
        
        \multirow{3}{*}{\textbf{YFCC4k} \cite{vo_et_al_2017}} 
        & GeoDecoder \cite{clark_et_al_2023} & $\sim 750$ & 10.3 & \textbf{24.4} & 33.9 & 50.0 & 68.7 \\
        & PIGEOTTO \textbf{[ME16]} & 418.8 & 9.5 & 22.5 & 38.8 & 60.7 & 76.9\\
        & PIGEOTTO \textbf{[ME16 + Landmarks]} & \textbf{383.0} & \textbf{10.4} & 23.7 & \textbf{40.6} & \textbf{62.2} & \textbf{77.7}\\

        \midrule

        \multirow{3}{*}{\textbf{YFCC26k} \cite{mueller_budack_et_al_2018}} 
        & GeoDecoder \cite{clark_et_al_2023} & $\sim 750$ & 10.1 & 23.9 & 34.1 & 49.6 & 69.0\\
        & PIGEOTTO \textbf{[ME16]} & 356.5 & 10.1 & 24.6 & 41.3 & 62.6 & 78.7\\
        & PIGEOTTO \textbf{[ME16 + Landmarks]} & \textbf{333.3} & \textbf{10.5} & \textbf{25.8} & \textbf{42.7} & \textbf{63.2} & \textbf{79.0}\\
        
        \midrule

        \multirow{3}{*}{\textbf{GWS15k} \cite{clark_et_al_2023}}
        & GeoDecoder \cite{clark_et_al_2023} & $\sim$ 2,500 & \textbf{0.7} & 1.5 & 8.7 & 26.9 & 50.5 \\
        & PIGEOTTO \textbf{[ME16]} & 440.8 & 0.1 & 8.7 & 30.1 & 64.0 & 84.7\\
        & PIGEOTTO \textbf{[ME16 + Landmarks]} & \textbf{415.4} & \textbf{0.7} & \textbf{9.2} & \textbf{31.2} & \textbf{65.7} & \textbf{85.1}\\
        \bottomrule
        \bottomrule
    \end{tabular}
    }
    \label{table:no_landmarks}
    %\vskip -0.08in
\end{table}

In \Cref{table:no_landmarks}, we observe that even when trained using the same data (ME16~\cite{larson_et_al_2017}), PIGEOTTO outperforms the prior SOTA Geodecoder~\cite{clark_et_al_2023} by a large margin on four out of five benchmarks. The improvements in benchmark results can largely be attributed to the end-to-end design of PIGEOTTO, not our final training data selection. Still, we find that including the 340,579 landmark images~\cite{weyand_et_al_2020} improves our model's performance across all benchmarks and distance metrics. We further note that both PIGEOTTO versions are also more robust than \citet{clark_et_al_2023}'s Geodecoder by almost an order of magnitude, reducing the median geolocalization error by more than $5x$ on the out-of-distribution (OOD) benchmark dataset GWS15k~\cite{clark_et_al_2023}. Given the benchmark and OOD results, PIGEOTTO is currently the only planet-scale image geolocalization model robust to location and image distribution shifts.

%-------------------------------------------------------------------------

\section{Auxiliary data sources}
\label{sec:datasets}

Our work relies on a wide range of auxiliary data that we can infer from each image's location metadata. This section details external datasets we are using either in the process of label creation or multi-task training.

\paragraph{Administrative area polygons.}

We obtain data on country areas from the \href{https://geodata.ucdavis.edu/gadm/gadm4.1/gadm_410-levels.zip}{Database of Global Administrative Areas} (GADM) \cite{gadm_2022}. Additionally, we obtain data on several granularities of political boundaries of administrative areas released by \href{https://github.com/wmgeolab}{The William \& Mary Geospatial Evaluation and Observation Lab} on GitHub. These data sources are used both in geocell label creation as well as to generate synthetic pretraining captions. The political boundaries are used in the semantic geocell creation process with Voronoi tesselations, as displayed in \Cref{fig:voronoi_tessellation}.

\paragraph{Köppen-Geiger climate zones.}

\begin{figure}[!htbp]
\centering
\vspace{10pt}
\includegraphics[width=1.0\textwidth]{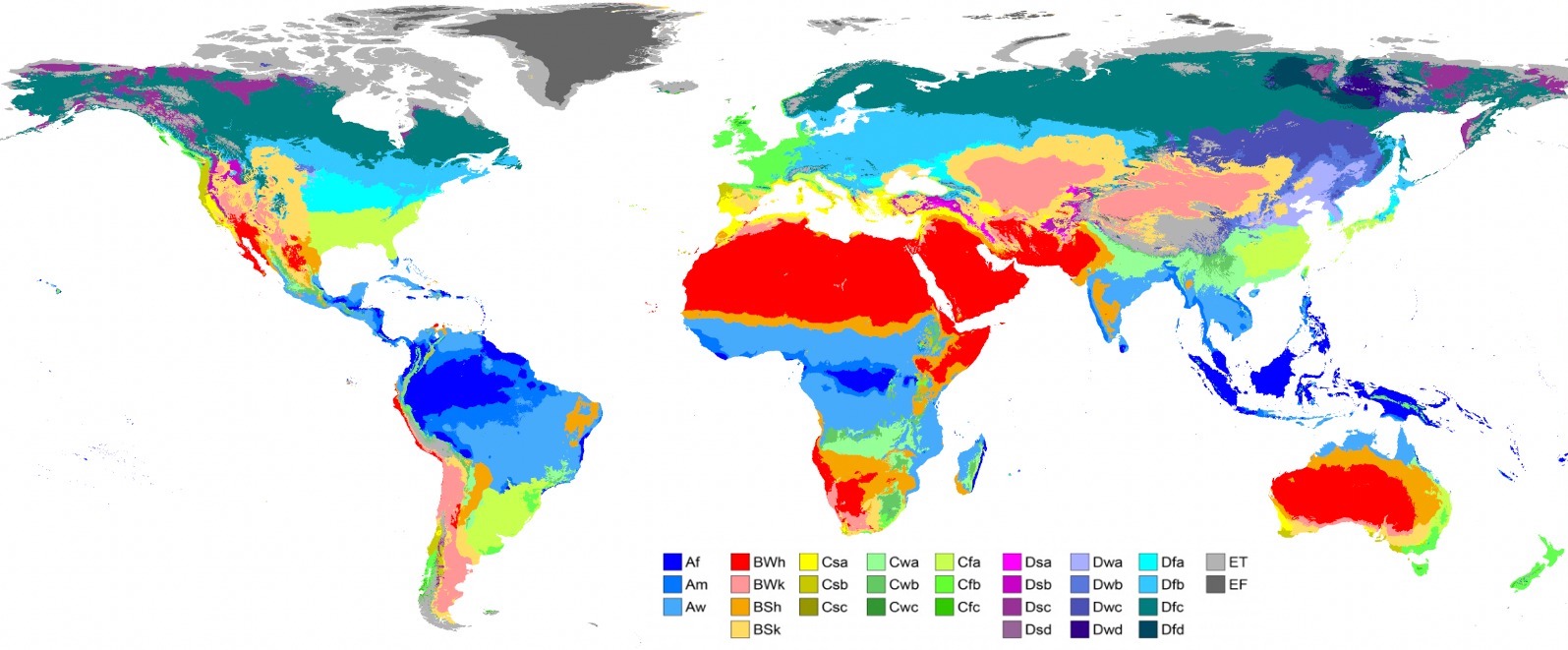}
\caption{Map of planet–scale Köppen-Geiger climate zones in our dataset. Adapted from \citet{beck_et_al_2018}.}
\label{fig:map_climate_zones}
\end{figure}

We obtain data on global climate zones through the \href{https://figshare.com/ndownloader/files/12407516}{Köppen-Geiger climate classification system} \cite{beck_et_al_2018}, visualized in \Cref{fig:map_climate_zones}. We use climate zone data both for synthetic caption generation for pretraining but also employ it in PIGEON's ablation study as a classification task (ablating "Multi-task Prediction Heads"), described in Tables \ref{table:ablation_study} and \ref{table:additional_results_distance}. The final PIGEON and PIGEOTTO versions only use climate zone data as part of their CLIP pretraining captions.

\paragraph{Elevation.}

We obtain data on elevation through the \href{https://www.usgs.gov/centers/eros/science/usgs-eros-archive-digital-elevation-shuttle-radar-topography-mission-srtm-1}{United States Geological Survey's Earth Resources Observation and Science (EROS) Center}. As elevation data was missing for several locations in our dataset, we augmented our data with missing values from parts of Alaska\footnote{\url{http://stacks.stanford.edu/file/druid:sg962yb7367/data.zip}} and Europe\footnote{\url{https://land.copernicus.eu/imagery-in-situ/eu-dem/eu-dem-v1.1/view}}. We use elevation data exclusively in a multi-task prediction setting via a log-transformed regression.

\paragraph{GHSL population density.}

We obtain data on population density through the \href{https://jeodpp.jrc.ec.europa.eu/ftp/jrc-opendata/GHSL/GHS_POP_GLOBE_R2022A/GHS_POP_E2020_GLOBE_R2022A_54009_1000/V1-0/GHS_POP_E2020_GLOBE_R2022A_54009_1000_V1_0.zip}{Global Human Settlement Layer} (GHSL). This data is also used in a multi-task prediction setting via a log-transformed regression.\pagebreak

\paragraph{WorldClim 2 temperature and precipitation.}

We obtain data on the average temperature, annual temperature range, average precipitation, and annual precipitation range through \href{https://www.worldclim.org/data/worldclim21.html}{WorldClim 2}. Similarly to prior auxiliary data, temperate and precipitation data is used in a multi-task regression setup, however, temperature values are not log-transformed before training.

\paragraph{Driving side of the road.}

We obtain data on the traffic direction through \href{https://www.worldstandards.eu/cars/list-of-left-driving-countries/}{WorldStandards}. This data is exclusively employed in generating synthetic pretraining captions.

%-------------------------------------------------------------------------
\section{Ablation studies on non-distance metrics}
\label{sec:ablation_studies}

Beyond the distance-based analysis of PIGEON described in the body of the paper, we also run ablation studies on non-distance metrics related to auxiliary data described in \Cref{sec:datasets}. In \Cref{table:ablation_study_non_distance}, we observe that our final PIGEON model version actually does not perform best on non-distance metrics related to a location's elevation, population density, season, and climate. The reason for this is that PIGEON does not share trainable model weights between the multi-task prediction heads and the location prediction tasks because joint multi-task training was already performed implicitly at the pretraining stage via synthetic captions. When sharing parameters between prediction heads (ablating ``Freezing Last Clip Layer"), a positive transfer between the tasks is observed and better performances are achieved on these auxiliary prediction tasks.\\

A key takeaway from \Cref{table:ablation_study_non_distance} remains that geographical, climate, demographic, and geological features can all be inferred from Street View images with potential applications in climate research and related fields.

\begin{table}[!htpb]
\centering
\caption{Results from the ablation study beyond the standard distance metrics, inferring geographical, climate, demographic, and geological labels from Street View imagery.}
\vspace{10pt}
\resizebox{1.0\textwidth}{!}{
\begin{tabular}{lcccccc}
\toprule
                                           & \textbf{Elevation} & \textbf{Pop. Density}              &   \textbf{Temp.}       &  \textbf{Precipitation}      & \textbf{Month}   & \textbf{Climate Zone}    \\
 \textbf{Ablation}                           & \textbf{Error}     & \textbf{Error}                     &   \textbf{Error}       &  \textbf{Error}              & \textbf{Accuracy} & \textbf{Accuracy}     \\
                                           & $m$          & $\nicefrac{people}{km^2}$   &   $^{\circ} C$ &  $\nicefrac{mm}{day}$ & $\%$       & $\%$           \\
\midrule
    \textbf{PIGEON}  & 149.6         & 1,119        & 1.26        & 15.08        & 45.42        & 75.22           \\
    \midrule
    $-$ Freezing Last CLIP Layer After Pretraining                  & \textbf{132.8}        & {1,072}        & \textbf{1.18}        & \textbf{12.82}        & \textbf{50.64}        & \textbf{75.76}          \\
    $-$ Contrastive CLIP Pretraining       & 147.1        & \textbf{1,064}     & 1.36       & 14.71      & 45.74      & 74.66      \\
    $-$ Semantic Geocells        & 141.7        & 1,094      & 1.37       & 14.48      & 45.74      & 74.10\\          
      
\bottomrule
\end{tabular}
}
\label{table:ablation_study_non_distance}
\end{table}

%-------------------------------------------------------------------------

\section{Additional analyses}
\label{sec:additional_analysis}

\subsection{Attention attribution examples}
\label{sec:attention_attribution_maps}

\begin{figure}[htbp]
\centering
\begin{subfigure}[b]{0.49\linewidth}
    \includegraphics[width=\columnwidth]{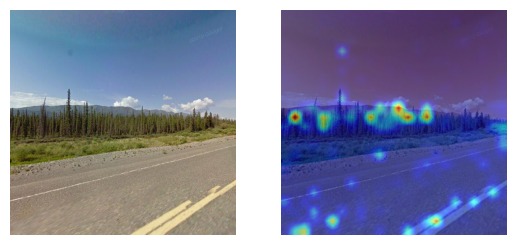}
    \caption{Attention attribution map for an image in Canada.}
    \label{fig:canada_attention_attribution_map}
\end{subfigure}
\hfill
\begin{subfigure}[b]{0.49\linewidth}
    \includegraphics[width=\columnwidth]{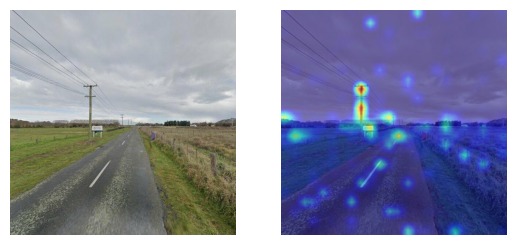}
    \caption{Attention attribution map for an image in New Zealand.}
    \label{fig:new_zealand_attention_attribution_map}
\end{subfigure}

\caption{Attention attribution maps for two locations in our Street View validation dataset.}
\vspace{10pt}
\label{fig:attention_attribution_maps}
\end{figure}

The contrastive pretraining used in CLIP gives the model a deeper semantic understanding of scenes and thereby enables it to discover strategies that are interpretable by humans. We observe that our model was able to learn strategies that are taught in online GeoGuessr guides without ever having been directly supervised to learn these strategies.

For the visualizations in Figure \ref{fig:attention_attribution_maps}, we generated attribution maps for images from the validation dataset and the corresponding ground-truth caption, e.g. ``This photo is located in Canada". Indeed, the model pays attention to features that professional GeoGuessr players consider important, like vegetation, road markings, utility posts, and signage, for example. This makes the strong performance of the model explainable and could furthermore enable the discovery of new strategies that professional players are not yet aware of.

\subsection{Urban vs. rural performance}
\label{sec:urban_vs_rural_performance}

\begin{figure}[!htbp]
\centering
\includegraphics[width=0.8\textwidth]{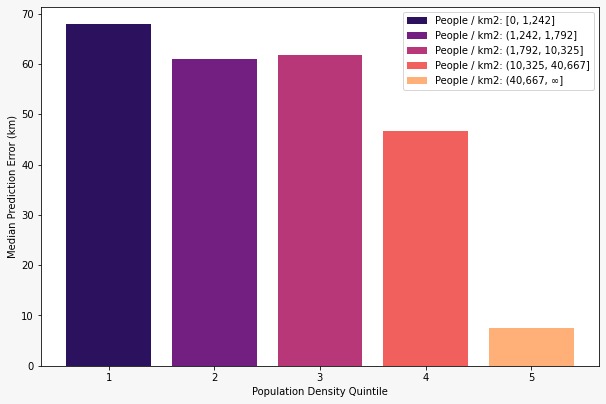}
\caption{Median km error by population density quintile.}
\label{fig:rural_vs_urban}
\end{figure}

In order to elucidate interesting patterns in our model's behavior, we investigate whether a performance differential exists for PIGEON in inferring the locations of urban versus rural images. Presumably, the density of relevant cues should be higher in Street View images from urban locations. Our analysis focuses on PIGEON because it has been trained on many rural images, whereas PIGEOTTO was trained predominantly on user-captured, urban images.\\

We bin our holdout Street View dataset into quintiles by population density and visualize PIGEON's median kilometer error. In Figure \ref{fig:rural_vs_urban}, we observe that higher population density indeed correlates with much more precise location predictions, reaching a median error of less than 10 kilometers for the 20 percent of locations with the highest population density.

\subsection{Qualitative analyses of failure cases}
\label{sec:failure_cases}

Despite our models' generally high accuracy in estimating image geolocations, there were several scenarios where they failed. We assess situations where our models were most uncertain and also identify the types of images for which our models made incorrect predictions.

\begin{figure}[!htb]
\centering

\begin{subfigure}[b]{0.49\linewidth}
    \includegraphics[width=\columnwidth]{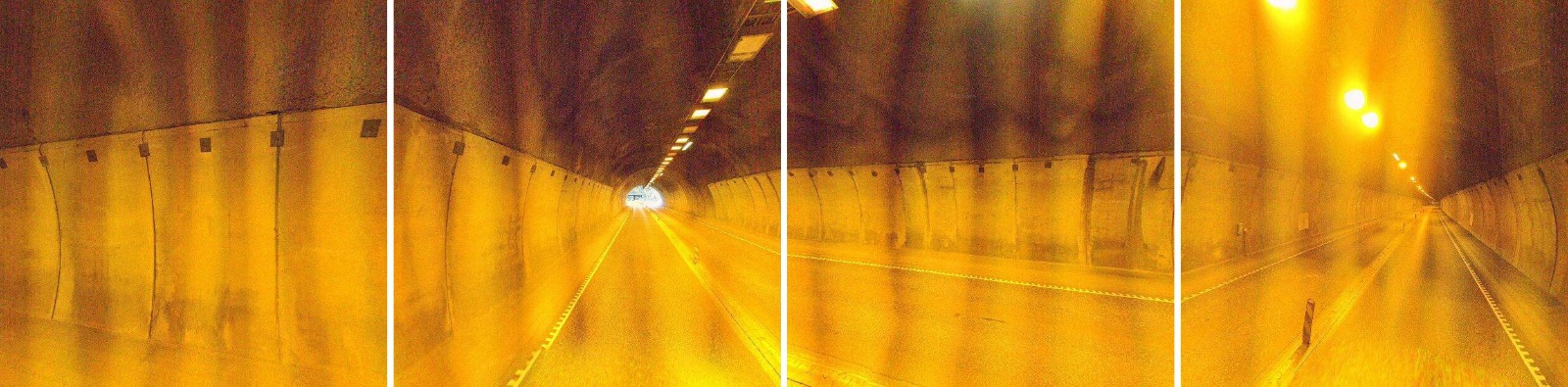}
    \caption{Image from a tunnel.}
    \label{fig:sub1}
\end{subfigure}
\hfill
\begin{subfigure}[b]{0.49\linewidth}
    \includegraphics[width=\columnwidth]{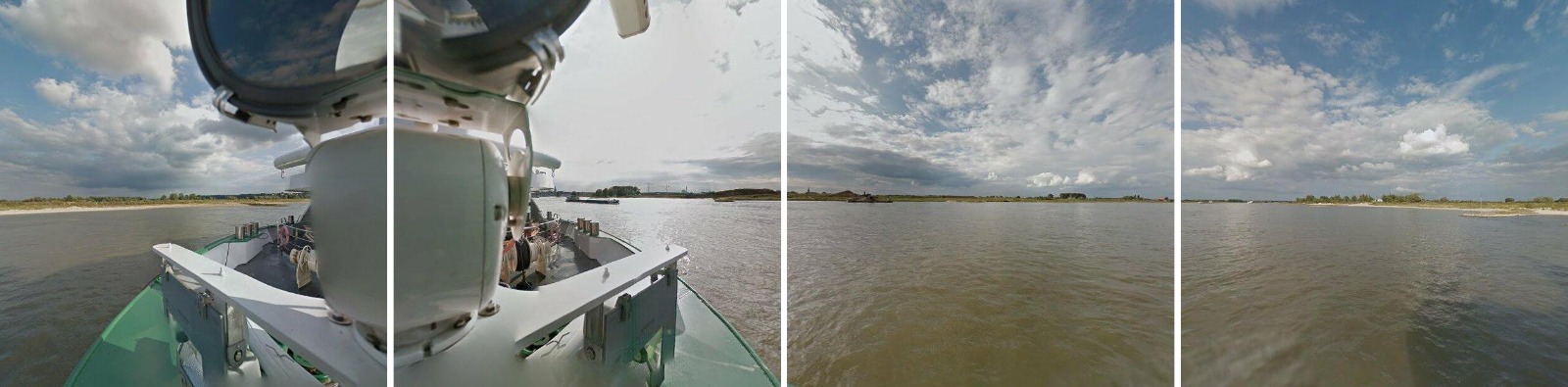}
    \caption{Image from a body of water.}
    \label{fig:sub2}
\end{subfigure}
\hfill
\begin{subfigure}[b]{0.49\linewidth}
    \includegraphics[width=\columnwidth]{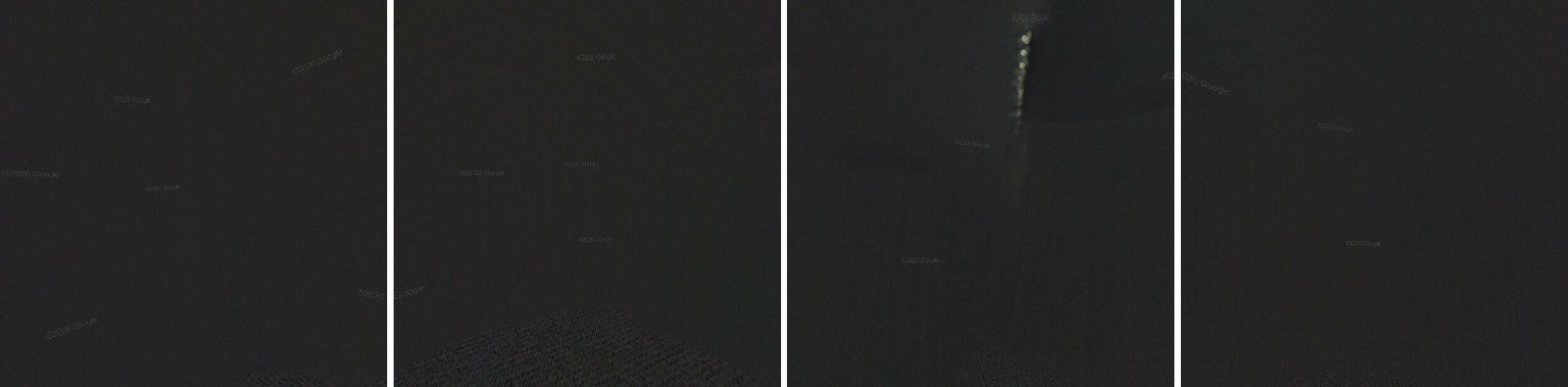}
    \caption{Image from a dark area.}
    \label{fig:sub3}
\end{subfigure}
\hfill
\begin{subfigure}[b]{0.49\linewidth}
    \includegraphics[width=\columnwidth]{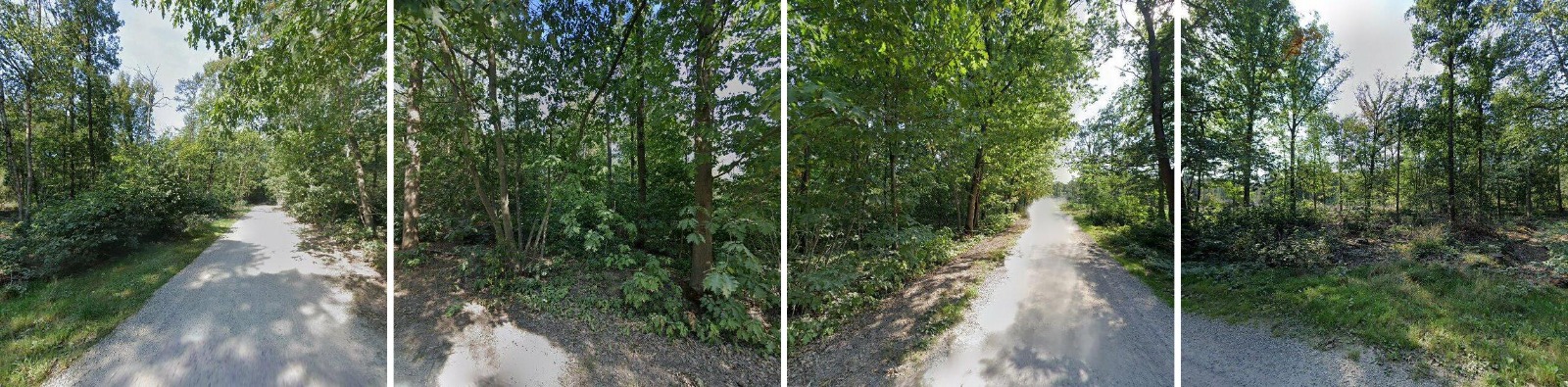}
    \caption{Image from a forest.}
    \label{fig:sub4}
\end{subfigure}
\hfill
\begin{subfigure}[b]{0.49\linewidth}
    \includegraphics[width=\columnwidth]{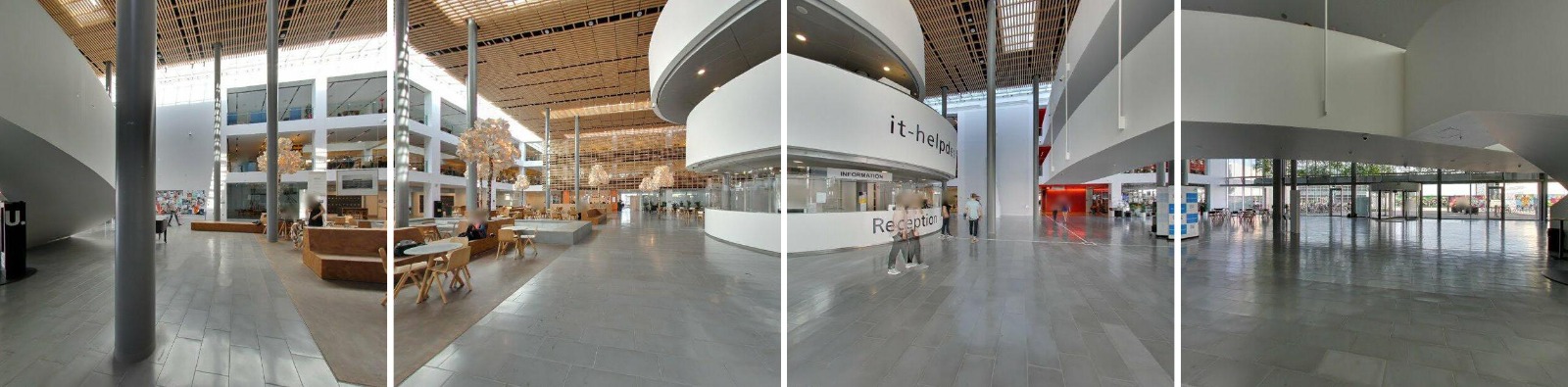}
    \caption{Image from an indoor area.}
    \label{fig:sub5}
\end{subfigure}
\hfill
\begin{subfigure}[b]{0.49\linewidth}
    \includegraphics[width=\columnwidth]{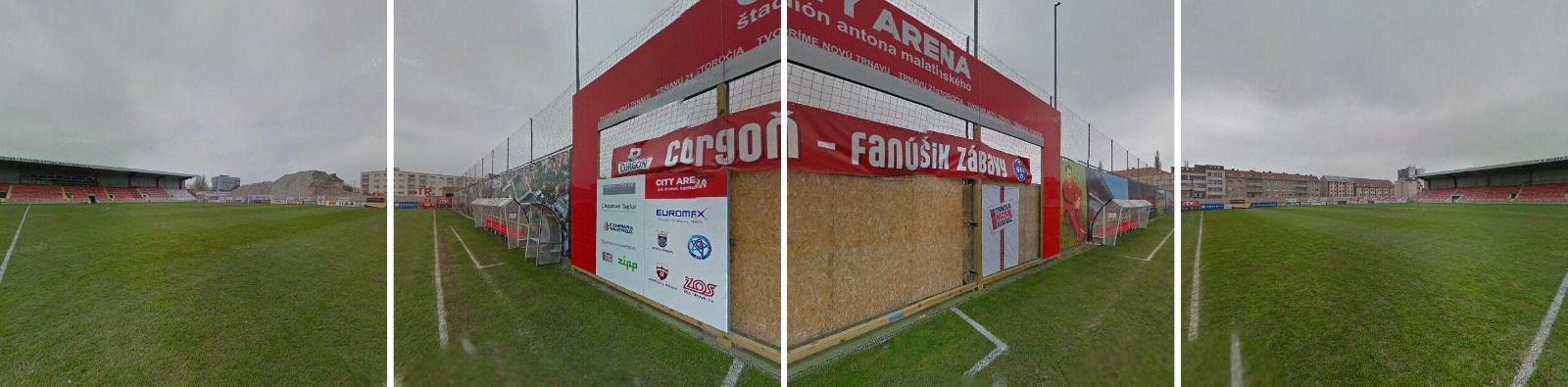}
    \caption{Image from a soccer stadium.}
    \label{fig:sub6}
\end{subfigure}
\caption{Examples of images for which PIGEON was the most uncertain about the correct location.}
\label{fig:failures}
\end{figure}

\paragraph{Uncertainty.} By computing the entropy over the probabilities of all geocells for each location in our validation set, we identified images where our models were most uncertain. For PIGEOTTO, these images were almost exclusively corrupted images remaining in the original Flickr corpus. For PIGEON, however, which was solely trained on Street View images, we observe some interesting failure cases in \Cref{fig:failures}. The features of poorly classified images are aligned with our intuitions and prior literature about difficult settings for image geolocation. Figure \ref{fig:failures} shows that images from tunnels, bodies of water, poorly illuminated areas, forests, indoor areas, and soccer stadiums are amongst the cases that are the most difficult to pinpoint by PIGEON.

\begin{figure*}[tbhp]
\vskip -0.2in
\begin{center}
\centerline{\includegraphics[width=0.78\linewidth]{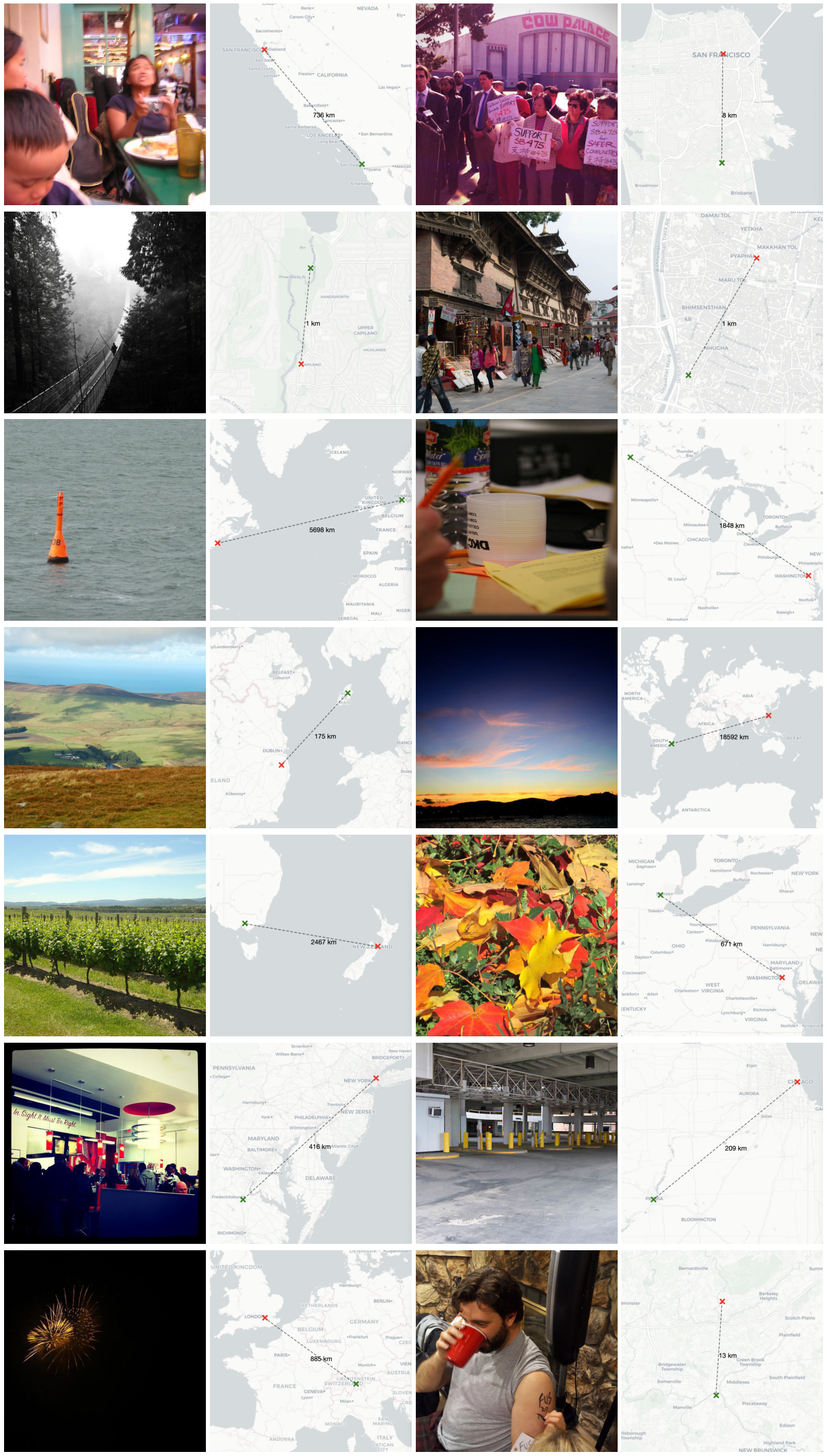}}
\caption{Example predictions of PIGEOTTO on fourteen images from the YFCC26k~\cite{mueller_budack_et_al_2018} test set.}
\label{fig:pigeotto_failure_cases}
\end{center}
\vskip -0.2in
\end{figure*}

\paragraph{Incorrect Predictions.} While for PIGEON, most failure cases are out-of-distribution (OOD) images that are atypical for Street View imagery (\Cref{fig:failures}), PIGEOTTO was trained to be highly robust to distribution shifts with the goal of being a general image geolocation system. To evaluate in what cases PIGEOTTO fails, we collected representative images from the YFCC26k~\cite{mueller_budack_et_al_2018} test set and plotted PIGEOTTO's predictions against the ground truth coordinates on a map in \Cref{fig:pigeotto_failure_cases}. We observe that real-world images (in this case derived from Flickr) are highly diverse, containing both indoor and outdoor images, vast differences in image depths, blurry images, images of people, filters that have been applied, and photos taken at night.\\

PIGEOTTO performs astoundingly well across a wide range of conditions; it correctly identifies popular places within circa one kilometer, as demonstrated by the third image in \Cref{fig:pigeotto_failure_cases} of the Capilano Suspension Bridge near Vancouver and the fourth image taken around the Kathmandu Durbar Square in Nepal. Our model fails in situations where the image contains very little information about the location, such as the fifth image containing only a boye in the sea, the sixth image containing a water bottle, or images taken at night with almost no visible features such as the eighth and thirteenth image, albeit in the later, PIGEOTTO does correctly predict Europe from the fireworks alone. PIGEOTTO further seems to work surprisingly well in indoor scenarios, even when images are blurred, as is the first image. Other examples include the eleventh image where our model still predicts the country correctly and the last image showing a person drinking from a red cup, guessed correctly to within 13 kilometers of the correct location. Finally, while PIGEOTTO confounds the wine regions of Victoria, Australia and Marlborough, New Zealand, it still is able to make accurate predictions from the flora alone, as evidenced by the close-up image of leaves, correctly guessed to within 671 kilometers.

\section{Deployment to GeoGuessr}
As part of our quantitative evaluation of PIGEON against human players, we develop a Chrome extension bot that uses PIGEON's coordinate output to directly place guesses within the game. This section is a high-level overview of our model serving pipeline.

\subsection{Data overlap in live games}

When deploying PIGEON in live games against human players, controlling for locations not in PIGEON's training dataset is impossible. Across all live games from \Cref{fig:geoguessr_players}, we find that $1.8\%$ of game locations are within less than $100$ meters of any location in our training dataset. This slight overlap between training and live evaluation locations is not problematic because top human players would see more unique locations over the course of their GeoGuessr career, resulting in an even larger overlap between already seen and new live data for them.

\subsection{Game mode}

GeoGuessr can be played in both single and multi-player modes. In our live performance evaluation of PIGEON, we decided to focus on GeoGuessr's \textit{Competitive Duels} mode, whereby the user directly competes with an opponent in a multi-round game with increasing round difficulty. Notably, while our GeoGuessr bot simply takes four images spanning the entire GeoGuessr panorama, other players can additionally move around in the Street View scene for at least 15 seconds which is the minimum time available to the opponent once a guess is made, resulting in them gathering more relevant information to refine their prediction. Each guess is subsequently translated into a GeoGuessr score whose formula we reverse-engineered by recording results from the game. The formula for the GeoGuessr score on the world map is approximately:

\begin{equation}\label{eq:geoguessr_score_function}
    \text{score}(x) = 5000 \cdot e^{-\frac{x}{1492.7}},
\end{equation}

where $x$ is the prediction error in kilometers.\\

\begin{figure}[!htbp]
\centering
    \begin{subfigure}{0.49\textwidth}
        \centering
        \includegraphics[width=\textwidth]{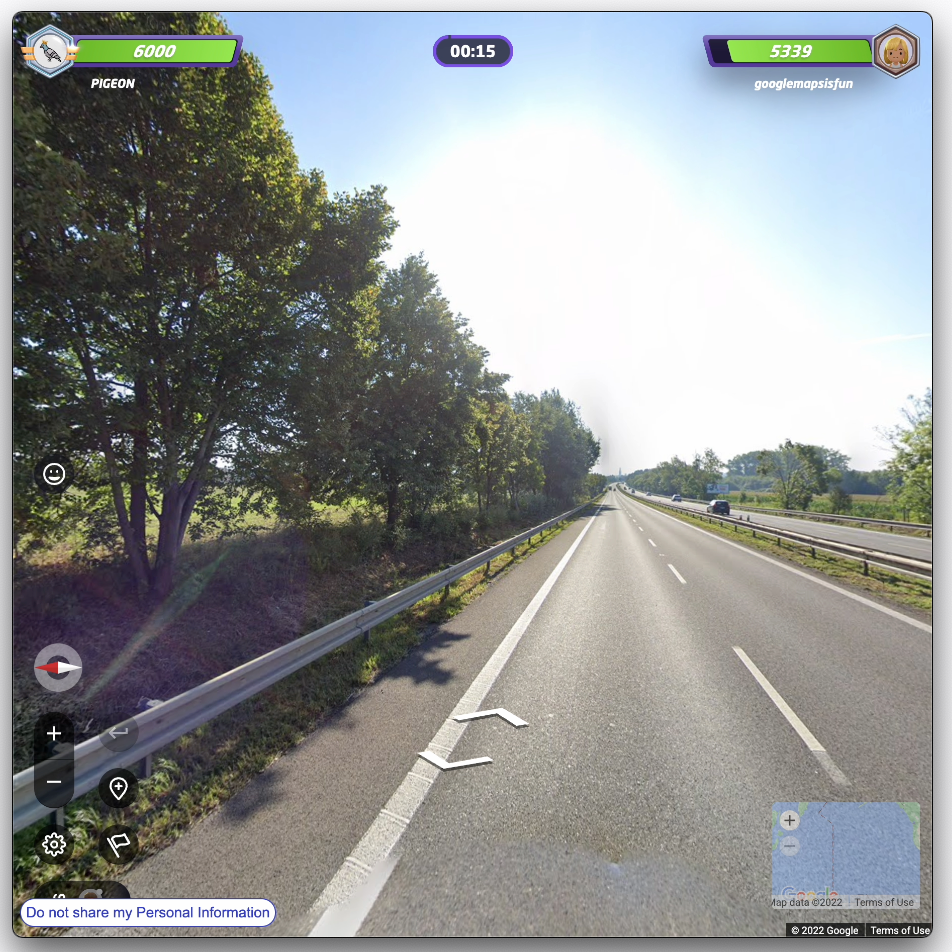}
        \caption{Sample image from a Geoguessr location.}
    \end{subfigure}
    \hfill
    \begin{subfigure}{0.49\textwidth}
        \centering
        \includegraphics[width=\textwidth]{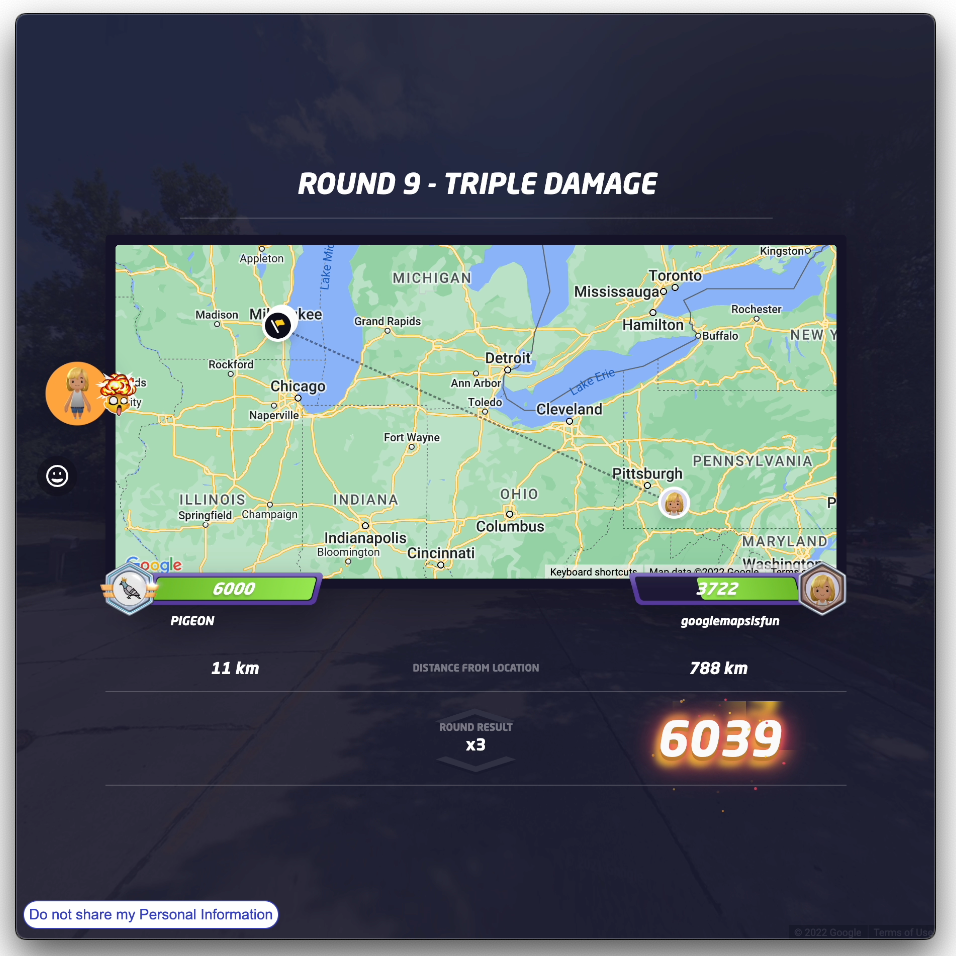}
        \caption{Comparison of a guess made by PIGEON and a human player.}
    \end{subfigure}
    \vspace{10pt}
    \caption{Sample screenshots from PIGEON deployed in the Geoguessr game.}
    \label{fig:geoguessr}
\end{figure}

To provide a better understanding of the Geoguessr game, \Cref{fig:geoguessr} shows two screenshots. The screenshots were taken while deploying PIGEON in-game against a human opponent in a blind experiment.\\

\subsection{Chrome extension}
We develop a GeoGuessr Chrome extension which is automatically activated once it detects that a game has started. It then autonomously places guesses in subsequent rounds, obtaining coordinate guesses from a PIGEON model API. The procedure to place a guess in the game works as follows and is repeated for each round until one player – PIGEON or its human opponent – has won:\pagebreak

\begin{enumerate}
    \item Resize the Chrome window to a square aspect ratio.
    \item Wait until the Street View scene is fully loaded.
    \item Repeat the following for all four cardinal directions:\\
    \begin{enumerate}
        \item Hide all UI elements.
        \item Take a screenshot.
        \item Unhide all UI elements.
        \item Rotate by $90^\circ$ using simulated clicks.\\
    \end{enumerate}
    
    \item Perform a POST request to our backend server with the four screenshots encoded as Base64 in the payload.
    \item Receive the predicted latitude \& longitude from our server.
    \item Optional: Random delay before making a guess to make the model's behaviour seem more human-like.
    \item Place a coordinate guess in the game by sending a request to GeoGuessr's internal API via the browser.
    \item Collect statistics about the true location \& human performance and submit them to the server using an additional POST request.
\end{enumerate}

\subsection{Inference API}
To serve image geolocalization predictions to our Chrome extension, we write code to serve PIGEON via an API on a remote machine with an A100 GPU. We utilize the Python library \href{https://fastapi.tiangolo.com/}{FastAPI} to implement two API endpoints:\\

\begin{itemize}
    \item \textbf{Inference endpoint.} A POST endpoint that receives either one or four images, passes them through a preprocessing pipeline and then runs inference on a GPU. In addition, it saves the images temporarily on disk for later evaluation. Finally, the API returns the latitude and longitude predictions of PIGEON to the client.\\
    \item \textbf{Statistics endpoint.} A POST endpoint that receives the statistics about the correct location, the score and distance of our guess, and human performance. This data is saved on disk and later used to generate summary statistics.\\
\end{itemize}

Our work demonstrates that PIGEON can effectively be applied in real-time scenarios as a system capable of end-to-end planet-scale image geolocalization.

\section{Acknowledgements}

We extend our gratitude to Erland Ranvinge, the Chief Technical Officer of GeoGuessr. Mr. Ranvinge's expertise and willingness to share in-depth knowledge about the GeoGuessr web API was instrumental in deploying PIGEON to GeoGuessr. His support in facilitating live evaluations against human players not only enhanced the quality of our  evaluations but also provided unique insights that enriched our findings. We further want to highlight that this work was initially conceived in an independent class project as part of \textit{CS 330: Deep Multi-Task and Meta Learning}, taught by Professor Finn at Stanford University, and subsequently evolved into the comprehensive study presented in this work. Finally, we wish to express our appreciation to Stanford's faculty and administrators who supported our work through cloud and compute resources.

\end{document}